\documentclass[10pt,journal,compsoc]{IEEEtran}
\ifCLASSOPTIONcompsoc
  \usepackage[nocompress]{cite}
\else
  \usepackage{cite}
\fi
\ifCLASSINFOpdf
  \usepackage[pdftex]{graphicx}
\else
  \usepackage[dvips]{graphicx}
\fi

\usepackage{amsmath}
\usepackage{amssymb}

\usepackage{tabularx}
\usepackage{multirow}

\usepackage{float}

\usepackage[dvipsnames]{xcolor}

\usepackage{caption}
\usepackage{subcaption}
\usepackage[most]{tcolorbox}
\usepackage{soul}

\usepackage[numbers]{natbib}
\begin{document}

% \title{Using knowledge distillation in Head Pose Estimation problem}
% \title{Knowledge distillation of offsets ensemble for the Head Pose Estimation problem}
% \title{Knowledge Distillation from Ensemble of Offsets for Head Pose Estimation}
% \title{Boosting of Head Pose Estimation by Knowledge Distillation from the Convolutional Ensemble}
\title{Boosting of Head Pose Estimation by Knowledge Distillation}


\author{Andrey~Sheka 
        and~Victor~Samun% <-this % stops a space
\IEEEcompsocitemizethanks{
\IEEEcompsocthanksitem A. Sheka and V. Samun are with the Krasovskii Institute of Mathematics and Mechanics, Ekaterinburg, Russia, 
and also with the Ural Federal University, Ekaterinburg, Russia.
\protect\\
E-mail: andrey.sheka@gmail.com and victor.samun@gmail.com}% <-this % stops a space
}

\IEEEtitleabstractindextext{%
\begin{abstract}
We propose a response-based method of knowledge distillation (KD) for the head pose estimation problem.
A student model trained by the proposed KD achieves results better than a teacher model, which is atypical for the response-based method.
Our method consists of two stages.
In the first stage, we trained the base neural network (NN), which has one regression head and four regression via classification (RvC) heads.
We build the convolutional ensemble over the base NN using offsets of face bounding boxes over a regular grid.
In the second stage, we perform KD from the convolutional ensemble into the final NN with one RvC head.
The KD improves the results by an average of 7.7\% compared to base NN.
This feature makes it possible to use KD as a booster and effectively train deeper NNs.
NNs trained by our KD method partially improved the state-of-the-art results.
KD-ResNet152 has the best results, and KD-ResNet18 has a better result on the AFLW2000 dataset than any previous method.
We have made publicly available trained NNs and face bounding boxes for the 300W-LP, AFLW, AFLW2000, and BIWI datasets.
Our method potentially can be effective for other regression problems.

\end{abstract}

}

\maketitle

% is not selected <OR> if conference mode is selected - because compsoc
% conference papers position the abstract like regular (non-compsoc)
\IEEEdisplaynontitleabstractindextext

% \fi
\IEEEpeerreviewmaketitle

\newcommand\optional[1]{}
\newcommand\optionalTwo[2]{#2}
\newcommand\optGreen[1]{#1}
\newcommand\optCyan[1]{#1}

\ifCLASSOPTIONcompsoc
\IEEEraisesectionheading{\section{Introduction}\label{sec:introduction}}
\else
\section{Introduction}
\label{sec:introduction}
\fi

% \IEEEPARstart{F}{acial} analysis в настоящее время является одним из важных направлений компьютерного зрения.
% \IEEEPARstart{F}{acial} analysis is one of the most important areas of computer vision. 
\IEEEPARstart{F}{acial} analysis is one of the most important tasks of computer vision. 
% Данное направление содержит такие задачи, как face detection \cite{Osadchy2007}, landmark detection \cite{Wu2019}, facial age estimation \cite{Chen2017}, head pose estimation \cite{Murphy-Chutorian2009} и другие \cite{Cao2018}.
This includes such tasks as face detection~\cite{Osadchy2007}, facial landmarks detection~\cite{Wu2019}, facial age estimation~\cite{Chen2017}, head pose estimation (HPE)~\cite{Murphy-Chutorian2009}, and others~\cite{Cao2018}.
% В данной работе мы фокусируемся на задаче head pose estimation.
In this paper, we focus on HPE.
% Ряд практических приложений, таких как facial expression analysis \cite{Zhang2018}, gaze estimation \cite{Langton2004}, driver monitoring system \cite{Liu2008,Murphy-Chutorian2010} и другие \cite{Marin-Jimenez2019,Passalis2020} требуют достаточно точного определения направления головы.
Many practical applications, such as analyzing facial expressions~\cite{Zhang2018}, gaze estimation~\cite{Langton2004}, \optionalTwo{driver monitoring systems~\cite{Liu2008,Murphy-Chutorian2010}, and others~\cite{Marin-Jimenez2019,Passalis2020}}{and others~\cite{Liu2008,Passalis2020}}, require fairly accurate HPE.

\begin{figure}[htbp]
    \centering
    \includegraphics[scale=0.8]{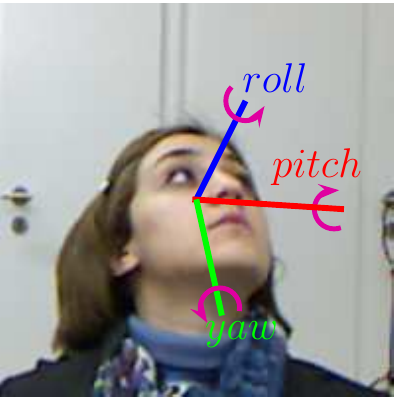}
    % \caption{Визуализация углов поворота головы}
    \caption{Visualization of head pose.}
    \label{fig_ang}
\end{figure}

% Направлением головы обычно называют вектор, содержащий углы pitch, yaw и roll.
The head pose is usually a vector containing the pitch, yaw, and roll angles.
% Данный вектор визуализируется с помощью трёх осей, как показано на рисунке~\ref{fig_ang}.
This vector is visualized using three axes, as shown in Fig.~\ref{fig_ang}.
% Задача head pose estimation заключается в нахождении данных углов по изображению лица.
The HPE problem is to find these angles from the image of the face.

% Для решения задачи head pose estimation было предложено множество подходов.
Many approaches have been proposed to solve the HPE problem.
% Одними из первых подходов были использование appearance template methods \cite{Huang2002}, деформируемых моделей \cite{Tzimiropoulos2017}, manifold embeddings methods \cite{Balasubramanian2007}, ключевых точек \cite{Zhu2016,Kumar2018} и других \cite{Hu2004,Huang2010}.
One of the first approaches was to use appearance template methods~\cite{Huang2002}, deformable models~\cite{Tzimiropoulos2017}, manifest embeddings methods~\cite{Balasubramanian2007}, facial landmarks~\cite{Zhu2016,Kumar2018}, and others~\cite{Hu2004,Huang2010}.
% Обзор данных методов приведён в работах \cite{Murphy-Chutorian2009,9291593}.
An overview of these methods is given in ~\cite{Murphy-Chutorian2009,9291593}.
% В последнее время для решения задач компьютерного зрения активно используются свёрточные нейронные сети \cite{Wang2019}.
% Recently, convolutional neural networks (NNs) have been actively used to solve many computer vision problems.
Recently, convolutional neural networks (NNs) have been widely used to solve many computer vision problems.
% Применительно к задаче head pose estimation, данный подход позволяет получить достаточно хорошие результаты \cite{9291593}.
Applying this approach to the HPE problem yields good results~\cite{9291593}.

% Типовой workflow для HPE problem на основе нейронных сетей содержит несколько операций. 
\optGreen{A typical workflow for the HPE problem based on NN contains several consecutive operations.
% Сначала детектор определяет ограничивающий прямоугольник вокруг лица.
First, the detector finds a bounding box around the face.
% Далее  изображение обрезается по найденному прямоугольнку. 
Next, the image is cropped according to the found bounding box and resized.
% После этого специализированная NN по обрезанному изображению определяет положение головы. 
After that, a specialized NN estimates the head pose from the cropped image.
}

% Усилия исследователей направлены на поиск способов кодирования углов, комбинаций head branch и функций потерь для улучшение HPE качества. 
% The research efforts aim to find ways to encode angles~\cite{Yang2019,Hsu2019,Cao2020}, combinations of head branches~\cite{Hsu2019,Wang2019,wang2019hybrid}, and loss functions~\cite{Ruiz2018,Wang2019,Zhang2020}, which improve HPE quality.
% Для улучшения качества нейронной сети исследователи  разработали различные способы кодирования углов, комбинаций head branch и функций потерь .
\optCyan{To improve the HPE quality of NNs, researchers have developed various methods to encode angles~\cite{Yang2019,Hsu2019,Cao2020}, configurations of head branches~\cite{Hsu2019,Wang2019,wang2019hybrid}, and loss functions~\cite{Ruiz2018,Wang2019,Zhang2020}.
% Однако,  датасеты используемые в тестовых протоколах имеют низкую репрезентативность.  
However, the datasets used in the test protocols have low representativeness.
% Мы показали~\cite{Sheka2021}, что аугментации вращения существенно улучшает репрезентативность обучающей выборки. 
% We have shown~\cite{Sheka2021} that rotation augmentation significantly improves the representativeness of the training sample.
% Простая нейронная сеть обученная с использованием аугментации вращения демострирует результат близкий к SOTA.
% Simple NN trained using rotation augmentation demonstrates a result close to complex SOTA NN~\cite{Sheka2021}.
% 
We have demonstrated that simple NN trained using rotation augmentation is close to the state-of-the-art (SOTA) result of complex NN~\cite{Sheka2021}.
In this paper, we propose an advanced augmentation workflow and a NN architecture with one regression head and four RvC heads to improve HPE accuracy.
}

% Одним из способов повышения точности предсказания также является обучение нескольких нейронных сетей с последующим усреднением их предсказаний \cite{Hansen1990,Sharkey1996,Yang2013}.
One way to additionally improve prediction accuracy is to train several NNs and then use an ensemble of their predictions~\cite{\optional{Hansen1990,}Sharkey1996,Yang2013}.
% Недостатком такого подхода является увеличение вычислительной сложности предсказания, т.к. необходимо получить предсказания от всех нейронных сетей и затем их усреднить.
The disadvantage of this approach is the increase in the computational complexity of the prediction since it is necessary to get predictions from all the NNs.
% Для решения данной проблемы в работах \cite{Liang2014,Yim2017} был предложен метод knowledge distillation.
The knowledge distillation (KD) method was proposed~\cite{Liang2014,Yim2017} to solve this problem.
% Суть метода заключается в обучении новой нейронной сети с использованием уже обученных сетей.
The essence of this method is to train a new NN using the trained NNs.
% В частности, для этого можно использовать обученный ансамбль.
% Данный метод применялся в ряде работ и показал довольно неплохие результаты \cite{Gou2020}.
This method has been used in several studies and has yielded good results~\cite{Gou2020}.
% Соответствующий обзор приведён в \cite{Gou2020}.
% Отметим также, что в работе \cite{Asif2020} показано, что использование knowledge distillation может повысить точность предсказания ансамбля.
% {\color{red} The paper~\cite{Asif2020} shows that KD can improve the accuracy of an ensemble prediction.}
% {
% \color{Green}
% The KD can improve the accuracy of an ensemble prediction~\cite{Asif2020}.
% }

% \color{Green}
% % В большинстве случаев модель ученика, обученная методом KD, показывает результаты хуже, чем модель учителя\cite{bla, bla2, bla3}.
% In most cases, the student model trained by the KD method shows worse results than the teacher model~\cite{10.1145/1150402.1150464, NIPS2014_ea8fcd92, hinton2015distilling}.
% % Хотя встречаются исключения, когда модель ученика может превзойти модель учителя~\cite{Asif2020}.
% There are exceptions when the student model is superior to the teacher model~\cite{Asif2020}.
% % В этих случаях KD может использоваться как бустинг для получения более точной модели.
% In these cases, the KD method can be used for boosting the accuracy of the teacher model.
% Простым и популярным способом KD является response-based метод, в котором ученик учится повторять response учителя.
\optCyan{A simple and popular way of KD is the response-based method, which trains the student to repeat the teacher response.
% Ученик, обученный response-based методом, всегда показывает результаты хуже учителя в задаче классификации~\cite{10.1145/1150402.1150464, NIPS2014_ea8fcd92, hinton2015distilling}.
A student trained by the response-based method always has results worse than the teacher model for the classification problem~\cite{Gou2020,10.1145/1150402.1150464, NIPS2014_ea8fcd92, hinton2015distilling}.
% Для задачи регрессии response-based методы оказываются неэффективными для NN with regression head. 
For the regression problem, response-based methods are ineffective for NN with a naive regression head.
The naive regression head response does not contain dark knowledge.
% эффективнее обучить нейронную сеть без учителя.
It is more efficient to train NN without a teacher.
}

% В работе \cite{Shao2019} показано, что для задачи head pose estimation на качество обученной нейронной сети существенное влияние оказывает bounding box.
In~\cite{Shao2019}, 
% {\color{red} it is shown }
\optGreen{\citeauthor{Shao2019} demonstrate} 
that the bounding box significantly affects the quality of the trained NN for the HPE problem.
% В работе \cite{Xue2020} предложен подход, при котором сначала производится корректировка bounding box и затем предсказание углов с помощью нейронной сети, представленной в предыдущей работе \cite{Yang2019},что позволило улучшить точность предсказания.
In~\cite{Xue2020}, 
% { \color{red} there is proposed} 
\optGreen{\citeauthor{Shao2019} propose} 
an approach in which the bounding box is first corrected, and then the angles are predicted using the NN  
% { \color{red}{presented in the previous paper~\cite{Yang2019}} }
% \optGreen{from previous work~\cite{Yang2019}.}
\optGreen{from~\cite{Yang2019}.}

 % Мы разработали сверточной ансамбль, который использует эту особенность ограничивающих прямоугольников.
\optCyan{We have developed a convolutional ensemble that uses this feature of bounding boxes.
% Сверточный ансамбль формирует пакет изображений, сдвигая исходную ограничивающую рамку поверх обычной сетки и обрезая изображение для каждой сдвинутой ограничительной рамки.
The proposed ensemble forms an image batch by shifting the original bounding box over the regular grid and cropping the image for each shifted bounding box. 
% Предсказанием сверточного ансамбля является усредненное предсказание нейронной сети по сформированному батчу изображений.
The ensemble prediction is the averaged NN predictions of obtained image batch.
% Для снижения вычистельных затрат мы применили KD из сверточного ансамбля.
We used KD from the convolutional ensemble to reduce computational costs.
% Неожиданно нейронная сеть, обученная с использованием дистилляции знаний, повысила точность определения углов поворота головы сверточного ансамбля.
Unexpectedly, NN trained using KD boosted the ensemble accuracy of HPE.
% При этом, что для задачи head pose estimation ранее не применялся подход knowledge distillation.
At the same time, KD has not yet been used for HPE.
}

% Подытоживая сказанное, перечислим цели данной работы:
This paper makes the following contributions.
\begin{itemize}
    \optCyan{\item 
% 	Мы предложили архитектуру нейронной сети с одной регрессионной головой и 4мя RvC головами. 
    % We have proposed a NN architecture with one regression head and 4 RvC heads.
% 		Для повышения репрезентативности данных мы использовали различные аугментации. 
    % We used various augmentations to increase the representativeness of the data.
% 		Предложенная архитектура и метод её обучения улучшают результаты многих предыдущих работ.
    % The proposed NN architecture and training method improve the results of many previous works.
    We propose a preprocessing workflow and a NN architecture with one regression head and four RvC heads.    
    Trained NN improves the results of many previous HPE works.
    
    \item
% 	- 	Мы предложили свёрточный ансамбль, который делает множественные обрезания картинки, используя смещения original face bounding box по регулярной сетке.
%     We have proposed a convolutional ensemble that makes multiple image cropping shifting the original bounding box over the regular grid. 
% % 		Ансамбль использует предложенную нейронную сеть для предсказания для каждого смещения прямоугольника.
%     The ensemble uses the proposed NN to predict each bounding box shift.
% % 		Предложенный ансамбль улучшает результаты нейронной сети.
%     The proposed ensemble improves the NN results.
    We propose a convolutional ensemble that makes multiple image cropping by shifting the original bounding box.
    The ensemble improves the NN results.
    \item
% %   Мы обучили NN используя KD из сверточному ансамбля.
%     We trained a NN using KD from a convolutional ensemble.
% % 		RvC голова позволила использовать метод переноса dark knowledge, разработанные для классификации, для регрессии. 
%     Using RvC head, we adapted for the regression problem the KD method developed for the classification problem.
% % 		Предоложенный KD workflow дал буст точности отностильно свёрточного ансамбля.
%     The proposed KD workflow boosted the convolutional ensemble accuracy.
% % 		NN обученная с помощью KD превосходит предыдущие работы на больших репрезентативных датасетах.
%     NN trained with KD surpasses previous results on large representative datasets.
    
    % Using RvC head, we adapted for the regression problem the KD method developed for the classification problem.
    % We adapt the classification KD method for the regression problem using RvC head.
    We propose the KD method for the regression problem using RvC head and classification KD loss.
    % The proposed KD workflow boosted the convolutional ensemble HPE accuracy.
    % The proposed KD workflow boosted the HPE accuracy of the convolutional ensemble.
    % NN trained with KD surpasses previous results on large datasets.
    NN trained with the proposed KD workflow boosts the convolutional ensemble accuracy and surpasses results on large datasets of previous HPE works.
    }
\end{itemize}

\section{Related work}

% В данном разделе мы дадим обзор подходам, применяемым для решения задачи head pose estimation.
This section gives an overview of the approaches used to solve the HPE problem. 
% В подразделе~\ref{relw_lbm} приведён обзор методов, основанных на использовании ключевых точек для определения углов поворота головы.
Subsection~\ref{relw_lbm} provides an overview of methods based on facial landmarks.
% В подразделе~\ref{relw_lfm} приведён обзор методов, не использующие ключевые точки для определения углов поворота.
Subsection~\ref{relw_lfm} describes an overview of landmark-free methods.
Subsection~\ref{relw_kd} contains a KD methods overview.

\subsection{Landmark-based methods} \label{relw_lbm}

% На первых порах для решения задачи head pose estimation применялись подходы с использованием ключевых точек.
% At first, approaches using facial landmarks were used to estimate the head pose.
Initially, facial landmark approaches were used for HPE.
% Подобные методы сначала  находят ключевые точки лица на изображении, используя, например, алгоритм из работы \cite{Wu2019}.
Such methods first find facial landmarks in the image, using, for example, the algorithm from~\cite{Wu2019}.
% Затем с использованием стандартной модели головы производится вычисление углов поворота головы, соответствующие алгоритмы приведены в \cite{Gao2003,Li2012}. 
Then the head pose is estimated using the average 3D head model and some algorithm~\cite{Gao2003,Li2012}. 
% Некоторые методы, например, \cite{Liu2021} в качестве дополнительной информации используют карту глубины, т.е. работают с 3D изображением.
Some methods, such as~\cite{Liu2021},  use a depth map as additional information.
% Применимость данных методов ограничена необходимостью использования специальных устройств, таких как RGB-D камер для получения изображений.
The applicability of these methods is limited by the need to use special devices to obtain RGB-D images.

% Однако, данные способы существенно зависят от точности нахождения ключевых точек и алгоритма расчёта углов.
% However, these methods significantly depend on finding facial landmarks and the algorithm for estimating head pose.
However, these methods significantly depend on finding facial landmarks.
% Если по тем или иным причинам некоторые ключевые точки трудно точно найти, например, часть лица перекрыта, точность вычисления углов ухудшится.
% If for some reason, some facial landmarks are difficult to find accurately, for example, part of the face is overlapped, the estimation of the head pose will deteriorate.
In some cases, facial landmarks are difficult to find accurately. 
For example, when part of the face is hidden.
% Одним из способов решения данных проблем может быть использование тепловых карт ключевых точек \cite{Gupta2019}.
One way to solve this problem could be to use heat maps of facial landmarks~\cite{Gupta2019}.
% Также может использоваться алгоритм RANSAC \cite{Fischler1981} или использование неполного набора ключевых точек \cite{Sheka2020}.
Also, the RANSAC~\cite{Fischler1981} algorithm can be used, or an incomplete set~\cite{Sheka2020} of facial landmarks can be used.
% Другой проблемой данных методов является использование усреднённой 3D-модели лица.
% Another problem with these methods is the use of an averaged 3D head model.
Another problem with these methods is using an averaged 3D head model.
% Отличие реальной формы анализируемого лица от реальной формы вносит внутреннюю ошибку (intrinsic error) в вычисления.
% Разница между фактической моделью 3D-головы и средней моделью 3D-головы вносит дополнительную ошибку в оценку.
The difference between the actual 3D head model and the average 3D head model introduces an additional error in the estimation.
% В работе \cite{Yuan2020} предложен способ решения данной проблемы. 
% {\color{red}The paper~\cite{Yuan2020} suggests a way to solve this problem.}
% \optGreen{In~\cite{Yuan2020}, \citeauthor{Yuan2020} suggest a way to solve this problem by using four facial landmarks and a 3D head model morphing with spherical parametrization.
% }
\optGreen{In~\cite{Yuan2020}, \citeauthor{Yuan2020} suggest solving this problem by using four facial landmarks and a 3D head model morphing with spherical parametrization.
}
% Однако, только при использовании информации о глубине получаются близкие к SOTA результаты.
% However, close to the \optionalTwo{state-of-the-art}{SOTA} results are obtained when only using depth information.
However, close to the \optionalTwo{state-of-the-art}{SOTA} results are obtained only using depth information.

\subsection{Landmark-free methods} \label{relw_lfm}

% В 2016 году в работах \cite{Ranjan2016,Chang2017,Gu2017} была продемонстрирована возможность решения задачи head pose estimation с помощью свёрточных нейронных сетей и без использования ключевых точек. 
In~\cite{Ranjan2016,Chang2017,Gu2017}, the possibility of solving the HPE problem using convolutional NNs without using key points is demonstrated.
% Данные подходы показывают более лучшие результаты, чем подходы, основанные на использовании ключевых точек. 
These approaches yield better results than approaches based on facial landmarks.
% Также данные методы менее подвержены проблемам, связанным с перекрытием части лица и менее зависят от точности нахождения ключевых точек.
Also, these methods are less susceptible to the problems associated with overlapping parts of the face, and they do not depend on the accuracy of finding facial landmarks.

% Одним из первых методов, показавших достаточно хорошие результаты, был HopeNet \cite{Ruiz2018}. 
% HopeNet~\cite{Ruiz2018} is one of the first methods that showed good results.
HopeNet~\cite{Ruiz2018} is one of the first methods to yield good results.
% Итоговое предсказание углов вычислялось с помощью регрессии через классификацию \cite{Torgo1996}. 
% The final prediction of the angles was calculated using Regression via Classification (RvC)~\cite{Torgo1996} with the bin size of 3. 
The final prediction of the angles was calculated using RvC~\cite{Torgo1996} with a bin size of 3. 
% Для классификации использовался размер бина 3. 
% RvC used the bin size of 3.
% Функция потерь в данном методе состояла из суммы классификации и регрессии. 
The loss function in this method combined classification and regression loss functions.
% В дальнейшем данный подход получил ряд улучшений. 
Subsequently, this approach received several improvements.
% Так, в работе \cite{Wang2019} была представлена модификация данного метода, в которой для итогового предсказания использовался размер бина 1, а функция потерь состояла из суммы для разных размеров бина. 
% {\color{red}The paper~\cite{Wang2019} presents}
\optGreen{In~\cite{Wang2019}, \citeauthor{Wang2019} present}
a modification of this method, in which the final prediction used an RvC head with a bin size of 1, and other RvC heads with bigger bin sizes were used as regularizers during the training. 
% В работе \cite{Huang2020} была представлена другая модификация метода, в которой предсказание вычислялось также с использованием регрессии через классификацию, но использовались топ-40 классов. 
% {\color{red}The paper~\cite{Huang2020} presented}
\optGreen{In~\cite{Huang2020}, \citeauthor{Huang2020} present}
another modification of the method. This modification also uses an RvC head, but the top 40 classes were used for the HPE. 
% Существует ряд работ \cite{Venturelli2017,Borghi2020}, в которых в качестве дополнительной информации используются карты глубины.
Some 
% {\color{red} papers} 
\optGreen{researchers}~\cite{Venturelli2017,Borghi2020} use depth maps as additional information.
% Несмотря на то, что данные работы показывают достаточно хорошие результаты, практическая применимость данных методов также ограничена наличием специальных камер.
Although these studies have obtained good results, the practical applicability of these methods is also limited by the availability of special cameras.

% {
% \color{red}
% % Одной из последних работ, превзошедшая все предыдущие, является \cite{Valle2020}.
% % Работа \cite{Valle2020} является одной из последних работ, превзошедшая все предыдущие результаты.
% % The paper~\cite{Valle2020} is one of the last papers that surpassed all previous results.

% The paper~\cite{Valle2020} is one of the latest papers to surpass all previous results.
% % Однако, при обучении данной сети используется информация о ключевых точках, которые, по сути, выступают регуляризатором для head pose estimation problem. 
% However, in that paper, when training an NN, information about facial landmarks is used, which is a regularizer for the HPE problem.
% % Авторы неявно расширили обучающий датасет с углами за счёт датасетов с ключевыми точками, что могло послужить причиной улучшения точности на углах. 
% They implicitly extended the training dataset with head poses by using datasets with facial landmarks, which could be the reason for their improvement of the accuracy of the HPE.
% }

% Одна из последних работ \cite{Valle2020} показала результаты, которые превосходят все предыдущие результаты.
\optGreen{One of the recent works~\cite{Valle2020} has results that surpass all previous results.
% Однако, при обучении данной сети используется информация о ключевых точках, которые, по сути, выступают регуляризатором для head pose estimation problem. 
% However, \citeauthor{Valle2020} in \cite{Valle2020} авторы используют информацию о ключевых точках, которые выступают регуляризатора для ead pose estimation problem. 
\citeauthor{Valle2020} in~\cite{Valle2020} extend the training dataset with head poses by a dataset with facial landmarks.
Authors train the NN to localize facial landmarks and estimate head pose.
The facial landmarks act as a regularizer.
The expansion of the dataset could significantly improve the representativeness of the data and lead to improving metrics for the HPE problem.
}

\subsection{Knowledge distillation methods} \label{relw_kd}

% Большие глубокие нейронные сети демонстрирует отличные результаты в задачах с большой обучающей выборкой.
\optCyan{Large deep NNs demonstrate excellent results on tasks with a large training sample.
% Избыточная параметаризация обеспечивает более тонкую настройку, что улучшает метрики качества в большинстве задач \cite{zhang2018informationtheoretic, pmlr-v97-brutzkus19b, 10.5555/3454287.3454840, 3e9e8ec9bde64cc295c1225f6cf98d23, Tu2020Understanding}.
Over parameterization provides fine-tuning, which improves quality metrics in most tasks~\cite{zhang2018informationtheoretic, \optional{pmlr-v97-brutzkus19b,}10.5555/3454287.3454840\optional{,Tu2020Understanding}}.
% \cite{zhang2018informationtheoretic, pmlr-v97-brutzkus19b, 10.5555/3454287.3454840, 3e9e8ec9bde64cc295c1225f6cf98d23, Tu2020Understanding}.
% \cite{Zhang et al., 2018; Brutzkus and Globerson, 2019; Allen-Zhu et al., 2019; Arora et al., 2018; Tu et al., 2020}.
% Однако, большие нейронные сети затруднительно использовать на мобильных устройствах и встраиваемых системах ввиду ограниченности вычислительных ресурсов.
% However, due to limited computing resources, large NNs are difficult to use on mobile devices and embedded systems.
However, large NNs are difficult to use on mobile devices and embedded systems due to limited computing resources.
% В некоторых случаях, маленькие нейронные сети обученные с нуля могут не обеспечивать требуемого качества. 
In some cases, small NNs trained from scratch may not provide the required quality.
% Данную проблему можно решать множеством методов: квантизация\cite{7780890}, прунинг\cite{248452}, дистиляция знаний\cite{hinton2015distilling}.
% This problem can be solved by a variety of methods: quantization~\cite{7780890}, pruning~\cite{248452}, \optionalTwo{knowledge distillation}{KD}~\cite{hinton2015distilling}.
This problem can be solved by various methods: quantization~\cite{7780890}, pruning~\cite{248452}, \optionalTwo{knowledge distillation}{KD}~\cite{hinton2015distilling}.
% Квантизация и прунинг сжимают большую модель, а дистиляция знаний обучает маленькую модель ученика с помощью большой модели учителя.
% Quantization and pruning compress a large model, and \optionalTwo{knowledge distillation}{KD} trains a small student model using predictions of a large teacher model.
Quantization and pruning compress a large model, and \optionalTwo{knowledge distillation}{KD} trains a small student model using large teacher model predictions.
% Данные методы в большинстве случаев могут использоваться совместно.
In most cases, these methods can be used together.
% Дистиляция знаний предоставляет гибкий набор параметров, что потенциально позволяет использовать данный метод не только в целях сжатия нейронных сетей~\cite{10.1145/1150402.1150464, NIPS2014_ea8fcd92, hinton2015distilling, 10.5555/2969033.2969123}
% ~\cite{Bucilua et al., 2006; Ba and Caruana, 2014; Hinton et al., 2015; Urban et al., 2017}.
% Дистиляция знаний имеет гибкую схему использования, что позволяет различных сценариев использования.
% Гибкая схема обучения дистилляции знаний позволяет настраивать её для различных сценариев.
A flexible training scheme of \optionalTwo{knowledge distillation}{KD} allows one to customize it for different scenarios.

% \cite{10.1145/1150402.1150464, NIPS2014_ea8fcd92, hinton2015distilling, 10.5555/2969033.2969123}
% \cite{7780890},  \cite{248452}, \cite{hinton2015distilling}

% Метод дистиляции знаний берёт своё начало в работе~\cite{10.1145/1150402.1150464}.
The \optionalTwo{knowledge distillation}{KD} method originates in~\cite{10.1145/1150402.1150464}. 
% Авторы в ~\cite{10.1145/1150402.1150464} предложили метод обучения маленькой модели используя результаты предсказания  большой модели или ансамбля моделей. 
\citeauthor{10.1145/1150402.1150464} in~\cite{10.1145/1150402.1150464} proposed a method of training a small model using the predictions of a large model or an ensemble of models.
% Позднее, Hinton et al., 2015 в~\cite{hinton2015distilling} формально популяризовал данный метод как дистиляции знаний.
Later, \citeauthor{hinton2015distilling} in~\cite{hinton2015distilling} formally popularized this method as \optionalTwo{knowledge distillation}{KD}.

% Изначально дистиляции знаний применялась для задач классификации.
Initially, \optionalTwo{knowledge distillation}{KD} was used for a classification problem.
% Предсказанное распределения вероятностей по классам является  dark knowledge модели учителя, которое помогает обучать модель ученика.
The predicted probability distribution of classes is the dark knowledge of the teacher model, which helps to train the student model.
% Такой метод дистиляции называется response-based, в котором ученик пытается полностью повторить поведение учителя.
This distillation method is called response-based since the student model tries to repeat the teacher model responses completely.
% Response-based подходы позволяют использовать в качестве учителя ансамбли без дополнительных доработок.
The response-based method allows using ensembles as a teacher without additional improvements.
% В response-based подходе для задачи классификации ученик всегда показывает результаты хуже учителя.
However, using this method, the student always has worse results than the teacher for a classification problem~\cite{Gou2020,10.1145/1150402.1150464, NIPS2014_ea8fcd92, hinton2015distilling}.

% Однако, во многих задачах использование response-based метода не показывает положительного эффекта, например,  для детекции.
% However, using the response-based method in many tasks does not show a positive effect.
However, using the response-based method in many tasks does not positively affect.
% В подобных задач применяется feature-based метод~\cite{}, в котором модель ученика пытается выучить ответ учителя на промежутчных слоях.    
For such tasks, the feature-based method trains the student model to predict the same features maps as the teacher model on the intermediate layers.
% Featrue-based метод не позволяет без дополнительных слоев-адаптеров использовать ансамбли в качестве учителя~\cite{}.
The feature-based method does not allow using ensembles as a teacher without additional adapter layers~\cite{park2019feed}.
% В некотороых случаях featrue-based метод позволяет модели ученика улучшить результат модели учителя. 
In some cases, the feature-based method allows the student model to improve the result of the teacher model~\cite{kim2020paraphrasing, \optional{10.1007/978-3-030-58595-2_40,}NEURIPS2020_657b96f0}.

% Примительно к задачи регрессии метод дистиляции знаний не применялся на основе response-based метода. 
The response-based method of \optionalTwo{knowledge distillation}{KD} has not been used for the regression problem.
% В большинстве случаев для регргессии используют наивное предсказание с помощью одного нейрона.
In most cases, naive coding is used, in which a single neuron predicts continuous values from a given range.
% Такое представление не позволяет модели учителя предоставить ученику уникальный dark knowledge.
This coding does not allow the teacher model to provide unique dark knowledge.
% Использование RvC головы позволяет адаптировать для задачи регргессии response-based методы, разработанные для задачи классификации.
% Using the RvC heads allows one to adapt for the regression problem to the response-based methods, which were developed for the classification problem.
Using the RvC heads allows one to adapt the response-based method to the regression problem. 
Initially, the response-based method was developed for the classification problem.
% Подобная схема дистиляции знаний для задачи регргессии ранее не применялась.
% This scheme of KD for the regression problem has not been used before.
The KD scheme using RvC heads has not been used before for the regression problem.

% ======== Кусок из обзора KD ==========
% ---- Пару предложений про большие модели, которые трудно использовать на устройствах. -----
% Large deep neural networks have achieved remarkable success with good performance, especially in the real-world scenarios with large-scale data, because the over parameterization improves the generalization performance when new data is considered (Zhang et al., 2018; Brutzkus and Globerson, 2019; Allen-Zhu et al., 2019; Arora et al., 2018; Tu et al., 2020). 
% However, the deployment of deep models in mobile devices and embedded systems is a great challenge, due to the limited computational capacity and memory of the devices. 
% % ---- Несколько предложений про историю KD -----
% To address this issue, Bucilua et al. (2006) first proposed model compression to transfer the  information from a large model or an ensemble of models  into training a small model without a significant drop  in  accuracy. 
% The learning of a small model from a  large  model  is later formally popularized as knowledge  distillation  (Hinton et al., 2015). 
% % ---- Схема teacher-student ------
% In knowledge distillation, a  small  student model is generally super-vised by a large teacher model (Bucilua et al., 2006; Ba and Caruana, 2014; Hinton et al., 2015; Urban et al., 2017). 

}

\section{Method}

% Идея предлагаемого подхода заключается в построении ансамбля смещений, который методом knowledge distillation обучает нейронную сеть, осуществляющей head pose estimation.
% The proposed approach aims to build an offsets ensemble that uses the KD method to train a NN that performs HPE.
% The approach proposed in the present paper aims to build an convolutional ensemble that uses the KD method to train an NN to then perform the HPE.
In this paper, we propose a neural network training method by KD from a convolutional ensemble for the HPE problem.
% Сначала мы в подразделе \ref{method_problem} сформулируем задачу head pose estimation.
First, we formulate the HPE problem in subsection~\ref{method_problem}.
% Затем, в подразделе \ref{method_arch} мы опишем архитектуры обучаемой нейронной сети.
Then, we describe the architectures of the trained NN in subsection~\ref{method_arch}.

% Получение итоговой нейронной сети будем выполнять в два этапа.
Getting the final NN is done in two stages.
% На первом этапе обучим нейронную сеть, решающую задачу head pose estimation.
% At the first stage, we train a NN that estimates head pose.
In the first stage, we train an NN to perform the HPE.
% На основе обученной модели нейронной сети построим ансамбль смещений, используя смещения bounding box.
We construct a convolutional ensemble based on the trained NN model and offsets of the bounding box.
% Соответствующий алгоритм приведён в подразделе \ref{method_ensemble}.
The algorithm is given in subsection~\ref{method_ensemble}.
% На втором этапе обучим итоговую нейронную сеть.
In the second stage, we train the final NN.
% При обучении используются результаты предсказания построенного ансамбля смещений.
The training uses the results of the prediction of the constructed convolutional ensemble.
% Подробное описание этапа приведём в подразделе \ref{method_distill}.
A detailed description of the KD is given in subsection~\ref{method_distill}.

% \subsection{Постановка задачи} \label{method_problem}
\subsection{Problem formulation} \label{method_problem}

% Задача head pose estimation состоит в нахождении углов pitch, yaw и roll для изображения с лицом.
The HPE problem is to use an image of a face to predict the pitch, yaw, and roll angles of that face.
% Пусть нам даны изображения $\mathcal{X} = \left\{X_1, \dots, X_N\right\}$ и pose vector $\vec{y}_k = \left(\alpha_k, \beta_k, \gamma_k\right)$ для каждого изображения $X_k$, где $N$ -- количество изображений, $\alpha_k$, $\beta_k$, $\gamma_k \in \left[-\theta; \theta\right]$ представляют углы pitch, yaw и roll соответственно.
Consider the images $\mathcal{X} = \left\{X_1, \dots, X_N\right\}$ and the pose vector $\vec{y}_k~=~\left(\alpha_k, \beta_k, \gamma_k\right)$ for each image $X_k$, where $N$ is the number of images, while $\alpha_k$,~$\beta_k$,~$\gamma_k \in \left[-\theta; \theta\right]$ represent the pitch, yaw, and roll angles, respectively.
% Целью является нахождение функции $\mathcal{F}$, предсказывающую углы поворота головы для изображения $X$, минимизировав mean average error~\eqref{eq_mae}. 
% The goal is to find a function $\mathcal{F}$ which estimates the head pose for the image $X$, minimizing the mean average error~\eqref{eq_mae}. 
The goal is to find a function $\mathcal{F}$ that estimates the head pose for the image $X$, minimizing the mean average error~\eqref{eq_mae}. 

\begin{equation}
    \label{eq_mae}
    E\left(\mathcal{X}\right) = \frac{1}{N} \sum_{i=1}^{N} \| \mathcal{F}\left(X_i\right) - \vec{y}_i \|_1
\end{equation}

% \subsection{Архитектура нейронной сети} \label{method_arch}
\subsection{Architecture of the neural network} \label{method_arch}

% Предложенная нами архитектура состоит из backbone и нескольких head branches.
\optCyan{Our proposed architecture consists of a backbone and several head branches.
% В качестве backbone используется нейронная сеть из семейства ResNet~cite{resnet}.
We use a NN from the ResNet family~\cite{He2016} as a backbone.
% На первой стадии обучения мы используем head branches состоящий из 1 регрессионная голова и 4 RvC головы с размерами бинов: 1, 2, 3, 4.
At the first stage of training, we use head branches consisting of one regression head and four RvC heads with bin sizes: 1, 2, 3, 4.
% На второй стадии обучения мы используем single head branch, которая является RvC голова с размером бина равным 1.
% In the second stage of training, we use a single head branch, which is an RvC head with a bin size of 1.
In the second stage of training, we use a single head branch, an RvC head with a bin size of 1.
% Архитектура нейронной сети представлена на рисунке .
% The architecture of the NN is shown in Fig.~\ref{fig_multihead}.
The NN architecture is shown in Fig.~\ref{fig_multihead}.

\begin{figure}[htbp]
    \centering
    \begin{minipage}[b]{\linewidth}
        \centering
        \includegraphics[width=\linewidth]{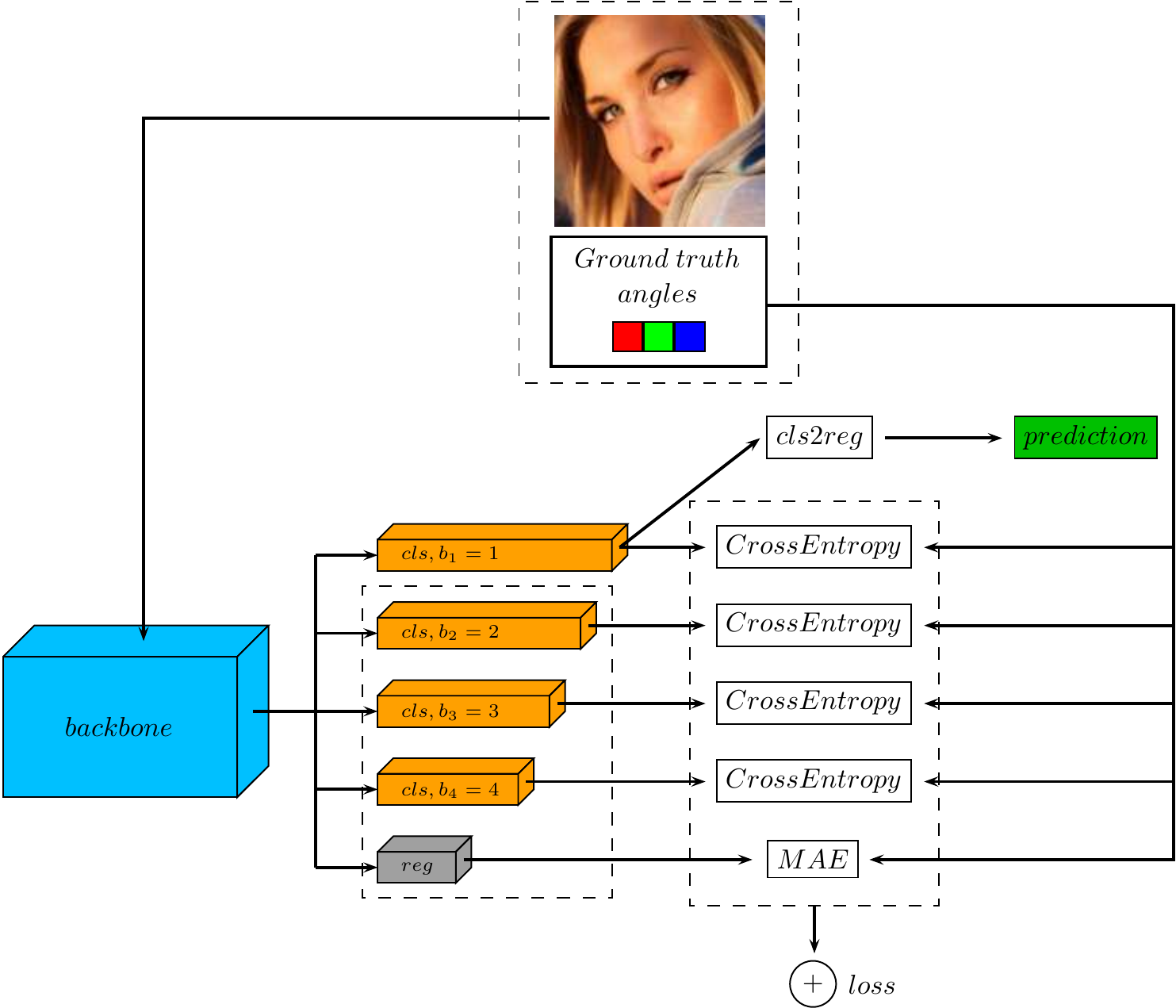}
        %\subcaption*{architecture}
        \vspace{.5em}
    \end{minipage}
    
    \begin{minipage}[b]{.45\linewidth}
        \centering
        \includegraphics[width=\textwidth]{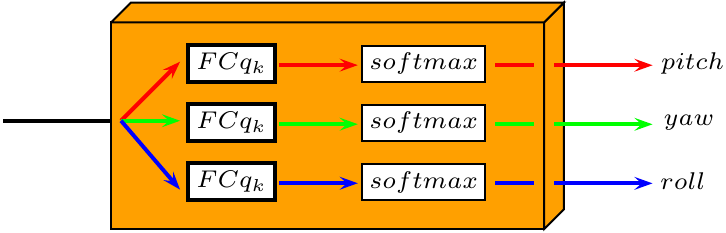}
        \subcaption*{classification head (cls)}
    \end{minipage}
    \hfill
    \begin{minipage}[b]{.45\linewidth}
        \centering
        \includegraphics[width=\textwidth]{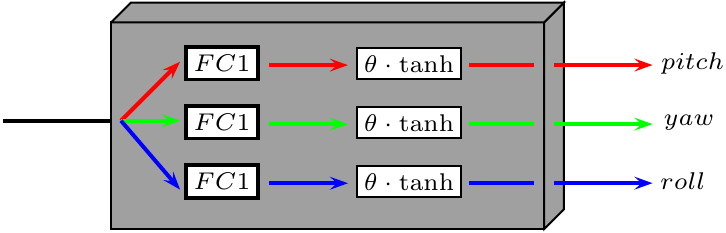}
        \subcaption*{regression head (reg)}
    \end{minipage}
    % \caption{Архитектура сети}
    \caption{
    Architecture of the neural network.
    \optGreen{ResNet* is used as a backbone.
    The NN has one regression head, and four RvC heads with bin sizes: 1, 2, 3, 4.
    The RvC head with bin size 1 is used for prediction.
    Other heads are used as regulators during the training.
    % В качестве backbone используется ResNet*. 
    % Нейронная сеть имеет одну голову нормированной до (-1,1) регресси и 4 RvC головы с размерами бинов: 1, 2, 3, 4.
    % Для предсказания используется RvC head c бин сайзом 1. 
    % Другие головы используются в качестве регулязаторов при обучении.
    % Головы предсказывать значение углов в диапазоне от -99 до 99.
    % 
    % На первой стадии используются все головы, на второй стадии регулязаторные головы удаляются.
    %
    }\optCyan{In the first stage, all heads are used. In the second stage, the regulatory heads are removed.
    }    
    }
    \label{fig_multihead}
\end{figure}

% В предыдущих исследованиях в качестве backbone использовалась ResNet50~\cite{}. 
Previous research~\cite{Ruiz2018,Wang2019,Shao2019,Huang2020} used ResNet50 as a backbone.
% В разделе экспериментов, мы показали, что использование более глубоких моделей ResNet101 и ResNet152  не позволяет повысить точность на первой стадии. 
% In the experiments section, we demonstrate that deeper ResNet101 and ResNet152 models do not improve accuracy significantly with single-stage training.
The experiments section demonstrates that deeper ResNet101 and ResNet152 models do not significantly improve accuracy with single-stage training.
% Это обосновывавает выбор ResNet50 as backbone в предыдущих работах~\cite{}.
% This justifies the choice of ResNet50 as a backbone in previous works~\cite{Ruiz2018,Wang2019,Shao2019,Huang2020}.
This result justifies the choice of ResNet50 as a backbone earlier.
% Однако, использование двух стадийного обучения позволяет получить прирост точности за счёт использования более глубоких моделей.
However, our two-stage training allows one to improve accuracy by using more deeper models.

% Предложенная нами конфигурация голов не использовалась ранее. 
The proposed head branches configuration has not been used before.
% В работе \cite{Ruiz2018}, \citeauthor{Ruiz2018} использовали одну RvC голову с joint classification/regression loss. 
\citeauthor{Ruiz2018} in \cite{Ruiz2018} used one RvC head with joint classification/regression loss.
% В работе \cite(wang2019hybrid), \citeauthor{wang2019hybrid} использовали несколько RvC голов с различными размерами бинов. 
\citeauthor{wang2019hybrid} in \cite{wang2019hybrid} used several RvC heads with different sizes of bins.
% Для головы с наименьшим размером бина использовали joint classification/regression loss.
The authors used joint classification/regression loss for the head with the smallest bin size.
% В работе \cite{Hsu2019}, \citeauthor{Hsu2019} исользовали одну голову регрессии и одну RvC, которая работала в режиме ранжирования.
\citeauthor{Hsu2019} in \cite{Hsu2019} used one regression head and one RvC head, which worked in ranking mode.
% Нами было установлено, что конфигурация из 4 голов RvC и 1 регрессионной обеспечивает наилучшее качество.
We found that the configuration with one regression head and four RvC heads provides the best quality.
% При этом удаление любой из них приводит к падению качества при обучении на 1ой стадии обучении.
At the same time, removing any of them leads to a drop in quality during training at the first stage.
% На второй стадии обучения достаточно одной RvC головы с размером бина равного единице.
At the second stage of training, one RvC head with a bin size equal to one is enough.
% Также использование joint classification/regression loss для RvC head не приводит к улучшению качества.
We also noticed that using joint classification/regression loss for RvC heads does not improve quality.
% Это может быть вызвано использование аугментаций, которые существенно повысили репрезентативность обучающей выборки.
This result may be caused by augmentations, which significantly increased the representativeness of the training sample.
% В нашей предыдущей работе было показано, что использование аугментации вращения и нейронной сети с одной RvC головой позволяет получить близкий к SOTA результат. 
% We showed~\cite{Sheka2021} that using rotation augmentation and a NN with one RvC head allows one to get a result close to the SOTA result.
% We showed~\cite{Sheka2021} that a single RvC head neural network has results close to SOTA.
% We showed~\cite{Sheka2021} that a single RvC head NN trained using rotation augmentation has results close to SOTA.
We demonstrated~\cite{Sheka2021} that a single RvC head NN trained using rotation augmentation achieves results close to SOTA.

% Одним из первых методов, показавших достаточно хорошие результаты, был HopeNet \cite{Ruiz2018}. 
% Итоговое предсказание углов вычислялось с помощью регрессии через классификацию \cite{Torgo1996}. 
% %
% В работе HOPE-net авторы использовали 4 головы RvC головы, но не использовали регрессионную голову.
% В работе ... авторы использовали 1 RvC голову и  
% Одним из первых методов, показавших достаточно хорошие результаты, был HopeNet \cite{Ruiz2018}. 
% Итоговое предсказание углов вычислялось с помощью регрессии через классификацию \cite{Torgo1996}. 
% Для классификации использовался размер бина 3. 
% Функция потерь в данном методе состояла из суммы классификации и регрессии. 
% В дальнейшем данный подход получил ряд улучшений. 
% Так, в работе \cite{Wang2019} была представлена модификация данного метода, в которой для итогового предсказания использовался размер бина 1, а функция потерь состояла из суммы для разных размеров бина. 
% В работе \cite{Huang2020} была представлена другая модификация метода, в которой предсказание вычислялось также с использованием регрессии через классификацию, но использовались топ-40 классов. 
% \cite{Ruiz2018} [HopeNet] Fine-Grained Head Pose Estimation Without Keypoints
% \cite(wang2019hybrid) [Hybrid] Hybrid coarse-fine classification for head pose estimation
% \cite{Shao2019} [TopK] Improving head pose estimation using two-stage ensembles with top-k regression
% \cite{Hsu2019} [QuatNet] QuatNet: Quaternion-based Head Pose Estimation with Multi-regression Loss
% 

% Далее мы приведем формальное описания голов в обобщенном виде.
Below we provide a formal description of the heads in a generalized form.
% На первом этапе нейронная сеть содержит несколько head branches: $M$ полносвязных слоёв, выполняющих классификацию с размерами бинов $B = \left\{b_1, \dots, b_M\right\}$, количеством бинов $Q = \left\{q_1, \dots, q_M\right\}$ и полносвязный слой, выполняющий регрессию.
% !!!!!!!!!!!!!!!!!!!!!!!!!!!!!!!!!!!!!!!!!!!!!!!!!!!!!
In the first stage, the NN contains several head branches: $M$ fully connected layers that perform the classification with bin sizes $B = \left\{b_1, \dots, b_M\right\}$, number of bins $Q = \left\{q_1, \dots, q_M\right\}$, and a fully connected layer performing a regression.
% !!!!!!!!!!!!!!!!!!!!!!!!!!!!!!!!!!!!!!!!!!!!!!!!!!!!!
% На втором этапе нейронная сеть содержит только один head branch: полносвязный слой, выполняющий классификацию с размером бина $b_k \in B$.
In the second stage, the NN contains only one head branch: a fully connected layer that performs a classification with the bin size $b_k \in B$.
}

\optGreen{Here we} describe the method for making the NN predictions during the training for the proposed architecture.
% Для классификации с размером бина $b_k$ и количеством бинов $q_k$ вычисление углов происходит по формуле~\eqref{calc_cls_output}, где $\vec{c} = \left(c_1, \dots, c_{q_k}\right) \in {\left[-\theta; \theta\right]}^{q_k}$ -- вектор центров бинов, $\vec{y\,}_{out}^{\left(k\right)} = \left(\vec{\alpha\,}_{out}^{\left(k\right)}, \vec{\beta\,}_{out}^{\left(k\right)}, \vec{\gamma\,}_{out}^{\left(k\right)}\right) \in \mathbb{R}^{q_k \times 3}$ -- выход $k$-й classification head. 
For a classification head with bin size $b_k$ and number of bins $q_k$, the angles are calculated using the formula~\eqref{calc_cls_output}, where $\vec{c} = \left(c_1, \dots, c_{q_k}\right) \in {\left[-\theta; \theta\right]}^{q_k}$ is the bin center vector, and $\vec{y\,}_\textup{out}^{\left(k\right)} = \left(\vec{\alpha\,}_{out}^{\left(k\right)}, \vec{\beta\,}_\textup{out}^{\left(k\right)}, \vec{\gamma\,}_\textup{out}^{\left(k\right)}\right) \in \mathbb{R}^{q_k \times 3}$ is the output of the $k$th classification head.
% Для регрессии вычисление углов происходит по формуле~\eqref{calc_reg_output}, где $\vec{y\,}_{out}^{reg} = \left(\alpha_{out}^{reg}, \beta_{out}^{reg}, \gamma_{out}^{reg}\right) \in \mathbb{R}^3$ -- выход regression head.
For a regression head, the angles are calculated using the formula~\eqref{calc_reg_output}, where $\vec{y\,}_\textup{out}^\textup{reg} = \left(\alpha_\textup{out}^\textup{reg}, \beta_\textup{out}^\textup{reg}, \gamma_\textup{out}^\textup{reg}\right) \in \mathbb{R}^3$ is the output of the regression head.

\begin{equation}
    \begin{aligned}
        \label{calc_cls_output}
        \xi_\textup{pred}^{\left(k\right)} = \langle \vec{c\,}, \operatorname{softmax} \left(\vec{\xi\,}_\textup{out}^{\left(k\right)}\right) \rangle && \text{where}~\xi \in \left\{\alpha, \beta, \gamma\right\}
    \end{aligned}
\end{equation}

\begin{equation}
    \begin{aligned}
        \label{calc_reg_output}
        \xi_\textup{pred}^\textup{reg} = \theta \tanh \left(\xi_\textup{out}^\textup{reg}\right) && \text{where}~\xi \in \left\{\alpha, \beta, \gamma\right\}
    \end{aligned}
\end{equation}

% В качестве функции потерь используется усреднённое значение функции потерь для $M$ классификаций и регрессии.
% The average of the values of the loss functions for the $M$ classification heads and the regression head is used as the loss function.
The resulting loss function is the averaging of the loss functions of all heads.
% Итоговая функция потерь $\mathcal{L}$ вычисляется по формулам~\eqref{multihead_loss}, где для некотрого изображения $\vec{y}_{gt} = \left(\alpha_{gt}, \beta_{gt}, \gamma_{gt}\right)$ -- ground truth pose vector, $\vec{y\,}_{out}^{\left(j\right)} = \left(\vec{\alpha\,}_{out}^{\left(j\right)}, \vec{\beta\,}_{out}^{\left(j\right)}, \vec{\gamma\,}_{out}^{\left(j\right)}\right)$ -- выход $j$-й classification head, $\vec{y\,}_{pred}^{reg} = \left(\alpha_{pred}^{reg}, \beta_{pred}^{reg}, \gamma_{pred}^{reg}\right)$ -- предсказание regression head. 
The resulting loss function $\mathcal{L}$ is calculated using the formulas~\eqref{multihead_loss}, where for some image $\vec{y}_\textup{gt} = \left(\alpha_\textup{gt}, \beta_\textup{gt}, \gamma_\textup{gt}\right)$ is the ground truth pose vector, $\vec{y\,}_\textup{out}^{\left(j\right)} = \left(\vec{\alpha\,}_\textup{out}^{\left(j\right)}, \vec{\beta\,}_\textup{out}^{\left(j\right)}, \vec{\gamma\,}_\textup{out}^{\left(j\right)}\right)$ is the output of the $k$th classification head, and $\vec{y\,}_\textup{pred}^\textup{reg} = \left(\alpha_\textup{pred}^\textup{reg}, \beta_\textup{pred}^\textup{reg}, \gamma_\textup{pred}^\textup{reg}\right)$ is the output of the regression head.

\begin{equation}
    %\small
    \begin{gathered}
        \mathcal{L}^{\left(j\right)}\left(\vec{y}_\textup{gt}, \vec{y\,}_\textup{out}^{\left(j\right)}\right) = \sum_{\xi \in \left\{\alpha, \beta, \gamma\right\}} \operatorname{CrossEntropy}\left(\xi_\textup{gt}, \vec{\xi\,}_\textup{out}^{\left(j\right)}\right) \\
        \mathcal{L}^\textup{reg}\left(\vec{y}_\textup{gt}, \vec{y\,}_\textup{pred}^\textup{reg}\right) = \operatorname{MAE}\left(\vec{y}_\textup{gt}, \vec{y\,}_\textup{pred}^\textup{reg}\right) \\
        \mathcal{L} = \frac{\mathcal{L}^\textup{reg}\left(\vec{y}_\textup{gt}, \vec{y\,}_\textup{pred}^\textup{reg}\right) + \sum_{j=1}^{M} \mathcal{L}^{\left(j\right)}\left(\vec{y}_\textup{gt}, \vec{y\,}_\textup{out}^{\left(j\right)}\right)}{M + 1}
    \end{gathered}
    \label{multihead_loss}
\end{equation}

% Для вычисления итогового предсказания нейронной сети используется выход полносвязного слоя, соответствующий классификации с размером бина $b_k$.
% Для вычисления предсказания нейронной сети в режиме тестирования используется выход полносвязного слоя классификации с минимальным размера бина.
The output of a fully connected classification layer with a minimum bin size $b_k$ is used to calculate the NN prediction in test mode.
% Остальные головы используются в качестве регуляризатора при обучении.
The remaining heads are used as regularizers during training.

% \subsection{Построение ансамбля} \label{method_ensemble}
\subsection{Convolutional ensemble} \label{method_ensemble}

% В работе \cite{Dietterich2000} показано, что одним из простых способов повышения точности алгоритмов машинного обучения является использование ансамбля. 
% {\color{red} The paper~\cite{Dietterich2000} shows }
\optGreen{In~\cite{Dietterich2000}, \citeauthor{Dietterich2000} demonstrates}
that one of the easiest ways to improve the accuracy of machine learning algorithms is to use an ensemble.
% Суть метода заключается в обучении нескольких моделей и последующего усреднения их предсказаний. 
This method consists of training several models and then averaging their predictions.
% В работах \cite{Shao2019,Xue2020} показано, что выбор bounding box влияет на результат предсказания. 
% {
% \color{red}
% The papers~\cite{Shao2019,Xue2020} demonstrate that the choice of bounding box affects the prediction result.
% }
\optGreen{Note that the choice of bounding box affects the prediction result~\cite{Shao2019,Xue2020}.
}

\begin{figure}[htbp]
    \centering
    \includegraphics[width=\linewidth]{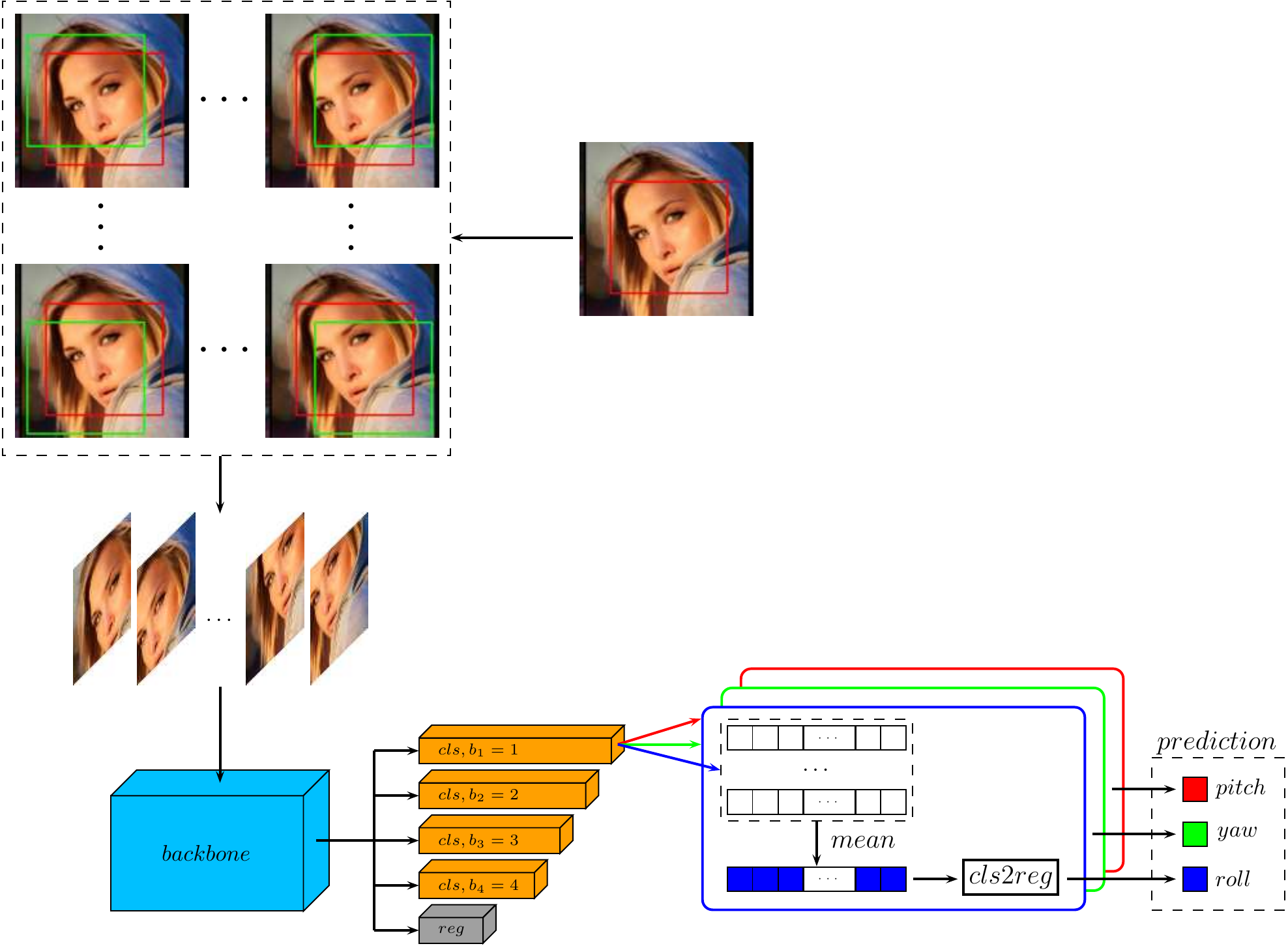}
    % \caption{Построение ансамбля}
    \caption{
    % {\color{red} 
    % Offset ensemble.
    % }
    \optGreen{Convolutional ensemble. 
    The ensemble forms an image batch by shifting the original bounding box over the regular grid.
    The red box indicates the original bounding box, the green box indicates shifted bounding box.
    The dimensions of the bounding box remain the same during offset.
    The NN predicts head poses for the batch.
    The final result of the ensemble prediction is the average result for the batch.
    % Ансамбль формирует пакет изображений смещая исходный bounding box по regular grid like convolution.
    % Нейронная сеть предсказывает head poses для всего пакета.
    % Итоговым результатом предсказания является усредний результат по пакету смещений. 
    }    
    }
    \label{fig_ensemble}
\end{figure}

% На основе этих фактов, у нас возникла идея использовать предсказания одной нейронной сети, но с разными смещениями прямоугольника.
\optCyan{Based on these facts, we decided to use the predictions of the single NN, but with different face bounding box offsets.

% Первоначально мы обучили нейронную сеть, чтобы получить предсказания нейронной сети в режиме квази-ансамбля для разных смещений ограничивающего прямогульника.
Initially, we trained the NN to obtain quasi-ensemble predictions for different offsets of the bounding box.
% Режим квази-ансамбля реализуется путём расширения входного обрезанного изображения.
% Для нейронной сети мы реализовали квази-ансамбль путём расширения входного изображения.
For a NN, we implemented a quasi-ensemble by expanding the input image.
% Если представить, что нейронная сеть это свертка с ядром 224x224, то она будет возвращать 1x1 предсказание углов для изображения с разрешением 224x224. 
Suppose that a NN is a convolution with a $224 \!\times\! 224$ kernel.
Then it returns a $1 \!\times\! 1$ prediction of angles for an image with a resolution of $224 \!\times\! 224$.
% Для изображения с разрешением 226x226 нейронная сеть соответственно вернет уже 3x3 значения предсказания, которые усредняются квази-ансамблем. 
For an image with a resolution of $226 \!\times\! 226$, the NN returns $3 \!\times\! 3$ predictions, which then are averaged by a quasi-ensemble.
% Отметим, что для увеличения разрешения необходимо также пропорционально увеличить обрамляющий прямоугольник, по которому производится обрезка изображения.
%Отметим, что для увеличения разрешения предварительно необходимо пропорционально увеличить обрамляющий прямоугольник. 
% Note that it is necessary to increase the bounding box proportionally to increase the input image resolution.
% Потому что входное изображение получается обрезкой изображения из датасета по прямоугольнику.
% Because the input image is obtained by cropping the dataset image along a bounding box.
It is necessary to increase the bounding box proportionally to increase the input image resolution because the input image is obtained by cropping the dataset image along a bounding box.
% Однако, точность предсказания квази-ансамбля падала более чем на 10\% по сравнению с базовым вариантом, предсказывающий одно значение без увеличения обрамляющего прямогульника.
% However, the HPE accuracy of the prediction of the quasi-ensemble decreased by more than 10\% compared to the variant without image extension.
However, the HPE accuracy of the quasi-ensemble decreased by more than 10\% compared to the variant without image extension.
% В режиме квази-ансамбля нейронная сеть не получает важную информацию о границах изображения в виде нулевых значений за счёт использования функции padding.
% В квази-ансамбле нейронная сеть не получает важную информацию о границах изображения.
The NN does not receive critical information about the image borders in a quasi-ensemble.
% Функция padding в нейронной сети обозначает границы нулями.
The padding function in the NN denotes borders with zeros.
% Однако, если делать обрезку изображений для каждого варианта смещения прямоугольника и независимо предсказывать значения нейронной сетью, то в таком варианте усреднения предсказаний даёт прирост точности.
However, cropping images for each bounding box offset and independent prediction by the NN increases the HPE accuracy due to averaging.
% Этот вариант мы назвали сверточным ансамблем, потому что смещения прямоугольников осуществляются по регулярной сетке как в светрке, но предсказания нейронной сетью происходят независимо для каждого смещения.
% Мы назвали этот вариант сверточным ансамблем, потому что смещения ограничивающей рамки выполняются на регулярной сетке, как в свертке, но нейронная сеть предсказывает положение головы независимо для каждого смещения.
% We call this variant the convolutional ensemble because the bounding box offsets are carried out on a regular grid as a convolution, but the NN predicts head pose independently for each offset.
We call this variant the convolutional ensemble because the bounding box offsets are carried out on a regular grid as a convolution.
However, the NN predicts head pose independently for each offset.
% Визуализация ансамбля смещений приведена на рисунке~\ref{fig_ensemble}.
The convolutional ensemble is shown in Fig.~\ref{fig_ensemble}.
}

% {
% \color{red}
% % Для построения ансамбля предлагается выполнять небольшие смещения bounding box и использовать одну нейронную сеть для предсказания.
% We propose to perform small bounding box offsets and use a single prediction network to construct the ensemble.
% % Использование смещений bounding box позволяет избежать необходимости обучения нескольки нейронных сетей для ансамбля. 
% Using bounding box offsets avoids training multiple NNs for an ensemble.

% % Построение ансамбля смещений для задачи head pose estimation состоит из обучения базовой нейронной сети и построения data flow для смещений изображений, используемых непосредственного для предсказании.
% Constructing an ensemble of offsets for the HPE consists of training the base NN and constructing a data flow for the image offsets used for prediction.
% % Визуализация ансамбля смещений приведена на рисунке~\ref{fig_ensemble}.
% The offset ensemble is shown in Fig.~\ref{fig_ensemble}.
% }

\begin{figure*}[t!]
    \centering
    \includegraphics[width=\linewidth]{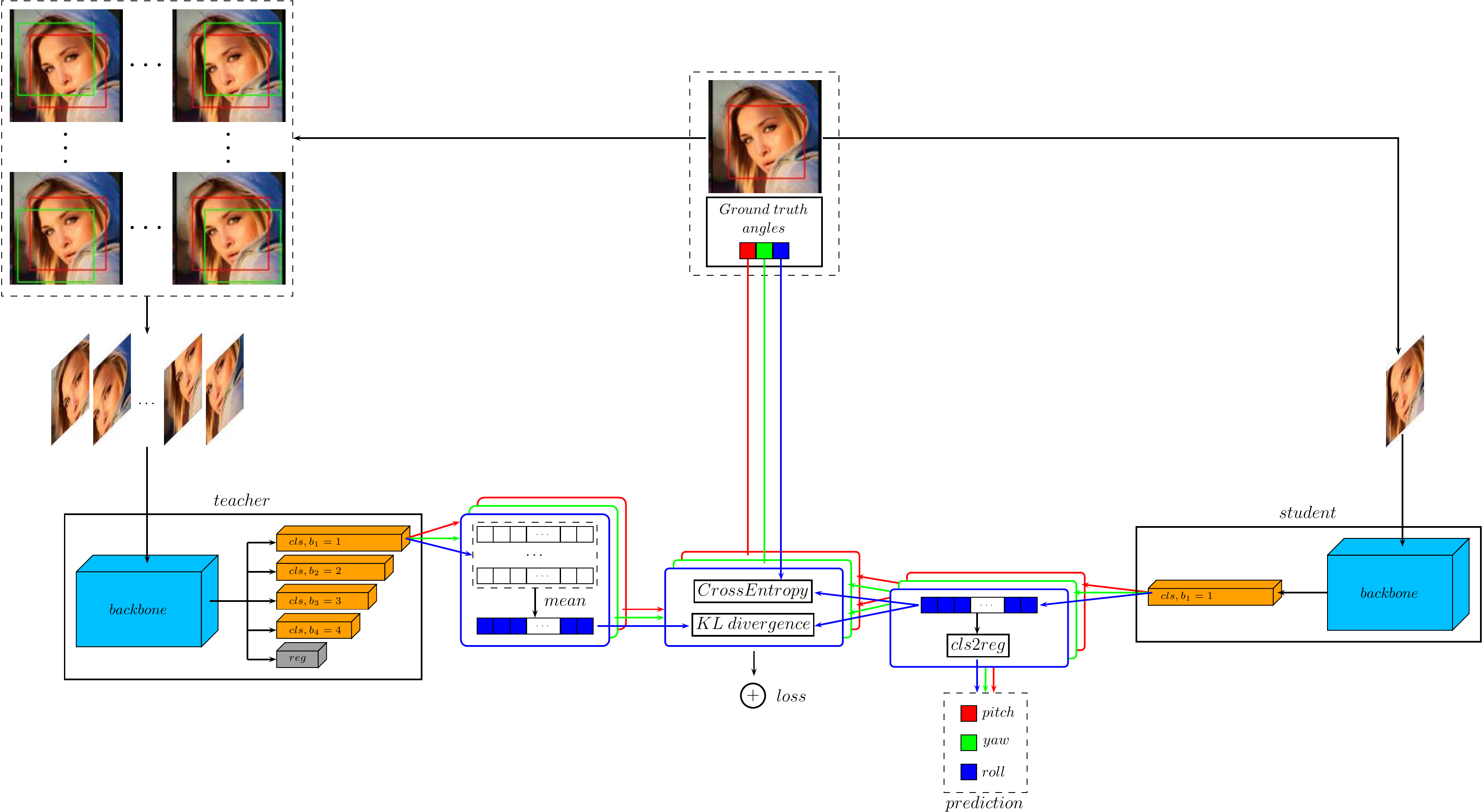}
    \caption{
    Knowledge distillation workflow.
    % Левая часть содержит модель учителя -- светрочный ансамбль, который по входному изображению оценивает положение головы.
    \optGreen{The left part contains the teacher model -- the convolutional ensemble that estimates the head pose from the input image.
    % Правая часть содержит модель ученика -- нейронную сеть с одно RvC головой, у которой размер бина равен 1.
    The right part contains the student model -- the NN with one RvC head, which bin size is 1.
    % Нейронная сеть ученика обучается на предсказаниях учителя и разметки из датасета.
    The student model is trained on the teacher model predictions and labeling from the dataset.
    }
    }
    \label{fig_kd}
\end{figure*}

% Сверточный ансамбль сделан на основе обученной нейронной сети, описанной в разделе ~\ref{method_arch}.
The convolutional ensemble is based on a trained NN described in section~\ref{method_arch}.
The NN has four RvC heads and one regression head.
% Для применения сверточного ансамбля необходимо предварительно подготовить изображение, обрезав лишние края и ресайзнуть его до нужных размеров. 
% Сверточный ансамбль предсказывает углы по подготовленному изображению с обрезанными краями и измененными размерами до требуемого размера.
% The convolutional ensemble predicts angles from the prepared image with cropped edges and resized to the required size.
The convolutional ensemble predicts angles from the prepared image, cropped by edges and resized to the required size.
% Полученное изображение фактически является receptive field сверточного ансамбля.
% The resulting image is the ensemble's receptive field.
The resulting image is the receptive field of the ensemble.
% Далее на основе множества смещений и полученного изображения получаем множество предсказаний нейронной сети.
% Затем мы вычисляем набор предсказаний нейронной сети, используя полученное изображение и набор смещений ограничивающей рамки.
Then, we calculate a set of NN predictions using the resulting image and a set of bounding box offsets.
% Полученные предсказния усредняются.
% Итоговым предсказанием ансамбля являются усредненное значение по множеству предсказаний.
The final prediction of the ensemble is the average value over the prediction set.

% Рассмотрим математическую формализацию сверточного ансамбля.
Consider the mathematical formalization of a convolutional ensemble.
% Введём следующие обозначения. 
Let introduce the following notation.
% Пусть параметр s и p - параметры ансамбля, где s - это stride, а p - это padding,a  входное изображение нейронной сети имеет размеры C x h x w.
Let $s \!\in\! \mathbb{N}$ and $p \!\in\! \left\{0, s, 2s, ...\right\}$ be the parameters of the ensemble, where $s$ is stride and $p$ is padding, and the input image of the NN has dimensions $C \!\times\! h \!\times\! w$.
% Для датасетного изображения X with bounding box с координатами x1, y1, x2, y2 необходимо применить сверточный ансамбль.
It is necessary to apply a convolutional ensemble for a dataset image $X \!\in\! \mathbb{R}^{C \!\times\! H \!\times\! W}$ with a bounding box having coordinates $\left(x_1, y_1, x_2, y_2\right)$.

% Давайте вычислим парамеры масштабирования датасетного изображения, используя формулы:
Let's calculate the scaling parameters of a dataset image using the formulas:
$r_x=\frac{w}{x_2 - x_1}$ and $r_y=\frac{h}{y_2 - y_1}$.
% и обрезки 
% Промасштабируем изображение с коэффициентами rx, ry.
Then, we scale the dataset image with the coefficients $\left(r_x, r_y\right)$.
% Далее обрежем scaled изображение прямоугольинком с координатами ...
% Then we crop the scaled image by box having coordinates $\left(x_1 r_x \!-\! p, y_1 r_y \!-\! p, x_2 r_x \!+\! p, y_2 r_y \!+\! p\right)$.
Then we crop the scaled image by $\left(x_1 r_x \!-\! p, y_1 r_y \!-\! p, x_2 r_x \!+\! p, y_2 r_y \!+\! p\right)$.
% В итоге получим входное изображение размера C x h+2p x w+2p для ансамбля.
% As a result, we get for the convolutional ensemble an input image $\widetilde{X}$ of size $C \!\times\! \left(h \!+\! 2p \right) \!\times\! \left(w \!+\! 2p\right)$.
As a result, we get an input  image~$Z \!\in\! \mathbb{R}^{C \!\times\! \left(h \!+\! 2p \right) \!\times\! \left(w \!+\! 2p\right)}$ for the convolutional ensemble.
% Отметим, что на данном этапе мы уже применили функцию padding, взяв недостающие значения из датасетного изображения.
Note that at this stage we have already applied the padding function, taking the missing values from the dataset image.

% Далее со stride получаем множество смещений .... . 
Next, we calculate the set of offsets coordinates $\mathcal{B}_{s,p} \left(w,h\right)$ by \eqref{offsets} using convolutional ensemble parameters $s$ and $p$. 
\begin{equation}
    \label{offsets}
    % \begin{aligned}
    %     % \mathcal{R}_{i,j} &= \left(is, js, w + is, h + js \right) \\
    %     % \mathcal{R}_{i,j} &= \left(si, sj, w + si, h + sj \right) \\
    %     \mathcal{R}_{i,j} &= \left(si, sj, si + w, sj + h  \right) \\
    %     % \mathcal{R}_{k,l} &= \left(ks, ls, w + ks, h + ls \right) \\
    %     % \mathcal{R}_{l, m} &= \left(ls, ms, w + ls, h + ms \right) \\
    %     % \mathcal{R}(k,l) &= \left(ks, ls, w + ks, h + ls \right) \\
    %     % O &= \left\{ \mathcal{R}_{i,j} \colon 0 \leq i, j \leq 2\frac{p}{s} \land i, j \in \mathbb{Z} \right\} \\
    %     O &= \left\{  \mathcal{R}_{i,j} \colon 0 \leq i, j \leq 2\frac{p}{s} \text{ and } i, j \in \mathbb{Z}  \right\} \\
    %     % O &= \left\{  \mathcal{R}_{l,m} \colon 0 \leq l, m \leq 2\frac{p}{s} \text{ and } l, m \in \mathbb{Z}  \right\} \\
    %     % O &= \left\{  \mathcal{R}(k,l) \colon 0 \leq i, j \leq 2\frac{p}{s} \text{ and } k, l \in \mathbb{Z}  \right\} \\
    % \end{aligned}
    \begin{aligned}
        \mathcal{R}_{i,j}^{\left(s\right)} \!\left(w,h\right) &= \left(si, sj, si + w, sj + h  \right) \\
        \mathcal{B}_{s,p} \left(w,h\right) &= \left\{  \mathcal{R}_{i,j}^{\left(s\right)} \!\left(w,h\right) \colon\! \left(0 \!\leq\! i,j \!\leq\! 2\frac{p}{s}\right) \!\land\! \left(i,j \!\in\! \mathbb{Z}\right) \right\} \\
    \end{aligned}
\end{equation}
% Для каждого смещения делаем кроп изображение и получаем предсказания нейронной сети.
For each $b \in \mathcal{B}_{s,p} \left(w,h\right)$, we crop the image $Z$ by $b$ and calculate the predictions of the NN.
% В итоге должно получиться ... предсказаний.
As the result we get $A = \left(2\frac{p}{s} + 1\right)^2$ predictions.
% Далее с помощью формулы 5 вычислем предсказание сверточного ансамбля. 
Next, we calculate the final prediction of the convolutional ensemble using formula~\eqref{ensemble}. 
% Отметим, что в данной формуле результатом являются вероятности.
Note that in this formula, the result is probabilities.

\begin{equation}
    \label{ensemble}
    \begin{aligned}
        \xi_\textup{ens}^{\left(j\right)} &= \frac{1}{A} \sum_{i=1}^{A} \left(\operatorname{softmax} \left(\vec{\xi\,}_\textup{out}^{\left(i\right)}\right)\right)_{j}, &\xi \in \left\{\alpha, \beta, \gamma\right\} \\
        \vec{\xi}_\textup{ens} &= \left(\xi_\textup{ens}^{\left(1\right)}, \dots, \xi_\textup{ens}^{\left(q_k\right)}\right), &\xi \in \left\{\alpha, \beta, \gamma\right\}
    \end{aligned}
\end{equation}

% Главное отличие сверточного ансамбля от сверточной нейронной сети заключается в обрезке изображения в каждом смещении и независимом предсказании углов для каждого смещения.
% В сверточной нейронной сети использование предсказаний соседних смещений, что позволяет сократить количество вычислений, но вносит дополнительный шум на границах каждого смещения.
% В сверточном ансамбле предсказания для каждого смещения происходят независимо,  что позволяет избежать шумов на границах изображения, но пропорционально количеству смещений увеличивает количество вычислений.
% Для нивелирования недостатка связанного с производительностью мы применили дистиляцию знаний.
\subsection{Knowledge distillation} \label{method_distill}

%\color{Cyan}
% Сверточный ансамбль использует множество предсказаний нейронной сети для одного изображения.
\optCyan{The convolutional ensemble uses multiple NN predictions for a single image.
% С одной стороны это повышает точность предсказания, но с другой стороны на порядок повышает количество вычислений.
On the one hand, this increases the accuracy of prediction, but on the other hand, it increases the number of calculations by orders of magnitude.
% Дополнительные вычисления снижают производительность и практическую применимость сверточного ансамбля.
Additional calculations reduce the performance and practical applicability of the convolutional ensemble.
% Для нивелирования недостатка снижения производительностью мы применили метод дистиляцию знаний.
We applied the KD method to neutralize the decreasing performance.

% \color{Red}
% В методе дистиляции знаний модель учителя обучает модели ученика.
The KD method train the student model to repeat the teacher model response.
A student trained by the response-based method always has results worse than the teacher model for the classification problem~\cite{Gou2020,10.1145/1150402.1150464, NIPS2014_ea8fcd92, hinton2015distilling}. 
For naive regression, the response-based method is useless.
% В нашем случае модель ученика превосходит модель учителя.
In our case, the student model is superior to the teacher model.
% В качестве учителя используется сверточный ансамбль из раздела~ref{}. 
% As a teacher model, we used a convlutional ensemble from the subsection~\ref{method_ensemble}.
As a teacher model, we used a convolutional ensemble.
% В качестве ученика используется нейронной сеть с одной RvC головой с размером бина 1 из раздела~ref{method_arch}. 
% As a student model, we used a NN described in the subsection~\ref{method_arch} with a single head branch, which is an RvC head with a bin size of 1. 
As a student model, we used a NN with a single head branch, which is an RvC head with a bin size of 1. 
% При обучении мы будем использовать метод knowledge distillation \cite{Liang2014,Asif2020}.
We train the NN using the KD method~\cite{Liang2014,Asif2020}, which is used for the classification problem.
% Визуализация процесса приведена на рисунке~\ref{fig_kd}.
The KD workflow is shown in Fig.~\ref{fig_kd}.
}

% {
% \color{Orange}
% % Напомним, что на данном этапе используется архитектура, описанная в подразделе \ref{method_arch}, но с одним head branch -- классификацией с размером бина $b_k \in B$.
% % Recall that at this stage, the architecture with a single head branch is used, described in the subsection~\ref{method_arch}.
% Recall that at this stage, the single head architecture described in subsection~\ref{method_arch} is used.
% This head branch is a classification head with a bin size of $b_k \in B$.
% % При обучении мы будем использовать метод knowledge distillation \cite{Liang2014,Asif2020}.
% We use the KD method~\cite{Liang2014,Asif2020} to train the NN.
% % В качестве учителя будут выступать предсказания по ансамблю нейронной сети, обученной на первом этапе. 
% The teacher is the offset ensemble.
% % Визуализация процесса приведена на рисунке~\ref{fig_kd}.
% The workflow of the KD is shown in Fig.~\ref{fig_kd}.
% }

% Функцию потерь будем вычислять по формулам~\eqref{dist_loss}, где $D_{KL}$ -- Kullback-Leibler divergence \cite{Kullback1951}, для некоторого изображения $\vec{y}_{gt} = \left(\alpha_{gt}, \beta_{gt}, \gamma_{gt}\right)$ -- ground truth pose vector, $\vec{y}_{out} = \left(\vec{\alpha}_{out}, \vec{\beta}_{out}, \vec{\gamma}_{out}^\right)$ -- выход обучаемой нейронной сети, $\vec{y}_{ens} = \left(\vec{\alpha}_{ens}, \vec{\beta}_{ens}, \vec{\gamma}_{ens}^\right)$ -- предсказание по ансамблю, вычисленное по формулам~\eqref{ensemble}.
The loss function \optCyan{for KD} is calculated using~\eqref{dist_loss}, where $D_\textup{KL}$ is the Kullback--Leibler divergence~\cite{Kullback1951}, for some image $\vec{y}_\textup{gt} = \left(\alpha_\textup{gt}, \beta_\textup{gt}, \gamma_\textup{gt}\right)$ is the ground truth pose vector, $\vec{y}_\textup{out} = \left(\vec{\alpha}_\textup{out}, \vec{\beta}_\textup{out}, \vec{\gamma}_\textup{out}\right)$ is the output of the trained NN, and $\vec{y}_\textup{ens} = \left(\vec{\alpha}_\textup{ens}, \vec{\beta}_\textup{ens}, \vec{\gamma}_\textup{ens}\right)$ is the prediction of the convolutional ensemble, calculated using~\eqref{ensemble}.

\begin{equation}
    \label{dist_loss}
    \begin{gathered}
        \mathcal{L}_\textup{cls}\left(\vec{y}_\textup{gt}, \vec{y}_\textup{out}\right) = \sum_{\xi \in \left\{\alpha, \beta, \gamma\right\}} \operatorname{CrossEntropy}\left(\xi_\textup{gt}, \vec{\xi}_\textup{out}\right) \\
        \mathcal{L}_\textup{dist}\left(\vec{y}_\textup{ens}, \vec{y}_\textup{out}\right) = \sum_{\xi \in \left\{\alpha, \beta, \gamma\right\}} D_\textup{KL}\left(\vec{\xi}_\textup{ens} \| \vec{\xi}_\textup{out}\right) \\
        \mathcal{L} = \frac{0.05 \cdot \mathcal{L}_\textup{cls}\left(\vec{y}_\textup{gt}, \vec{y}_\textup{out}\right) + \mathcal{L}_\textup{dist}\left(\vec{y}_\textup{ens}, \vec{y}_\textup{out}\right)}{2}
    \end{gathered}
\end{equation}

% Вычисление результатов предсказания выполняется по формулам~\eqref{calc_cls_output}, приведённым ранее.
The prediction is calculated using the following formulas~\eqref{calc_cls_output}.
% Отметим также, что на данном этапе мы можем выбрать другой backbone, отличный от используемого на первой стадии.
% Note that we can choose a different backbone from the one used in the first stage at this stage.
Note also that at this stage, we can choose a different backbone from the one used in the first stage.

\section{Datasets}

% В данном разделе мы приведём описание данных, используемых в экспериментах.
In this section, we describe the datasets used in the experiments.
% В подразделе~\ref{ds_desc} приведём обзор датасетов, используемых в задаче head pose estimation.
Subsection~\ref{ds_desc} provides an overview of the datasets used.
% Далее, в разделе~\ref{ds_prep} опишем препроцессинг данных.
Subsection~\ref{ds_prep} describes the data preprocessing.
% В разделе~\ref{ds_aug} приведём список используемых аугментаций.
Subsection~\ref{ds_aug} contains a list of the augmentations used.
% Наконец, в разделе~\ref{ds_proto} приведём используемые тестовые протоколы.
Lastly, subsection~\ref{ds_proto} describes the test protocols used for the HPE.

\subsection{Description} \label{ds_desc}

\begin{figure}[bh]
    \centering
    \includegraphics{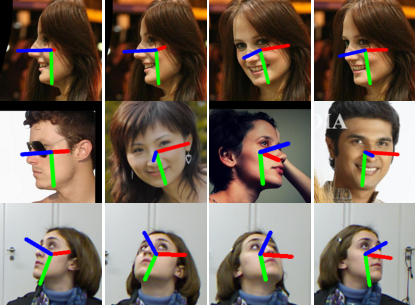}
    % \caption{Изображения из датасетов. \\
    % Первая строка содержит примеры из 300W-LP, вторая строка содержит примеры из AFLW2000, третья строка содержит примеры из BIWI}
    \caption{Examples of dataset images. 
    % \\
    The first line contains examples from 300W-LP, the second line contains examples from AFLW2000, and the third line contains examples from BIWI.
    % The first line contains examples from 300W-LP, the second line -- AFLW2000, the third line -- BIWI.
    %Each line contains examples from 300W-LP, AFLW2000, and BIWI, respectively.
    }
    \label{fig_datasets}
\end{figure}

% В работах, посвящёных задаче head pose estimation, обычно используются датасеты 300W-LP \cite{Zhu2016}, AFLW2000 \cite{Zhu2016}, AFLW \cite{Kostinger2011} и BIWI \cite{Fanelli2013}.
Research on HPE usually uses the datasets 300W-LP~\cite{Zhu2016}, AFLW2000~\cite{Zhu2016}, AFLW~\cite{Kostinger2011}, and BIWI~\cite{Fanelli2013}.
% Датасет 300W-LP синтетически сгенерирован из датасетов AFW~\cite{Zhu2012}, LFPW~\cite{Belhumeur2013}, HELEN~\cite{Zhou2013} и IBUG~\cite{Sagonas2013} с помощью деформируемых 3D-моделей. 
\optionalTwo{The 300W-LP dataset}{300W-LP} is synthetically generated using deformations of a 3D face model from \optional{the datasets} AFW~\cite{Zhu2012}, LFPW~\cite{Belhumeur2013}, HELEN~\cite{Zhou2013}, and IBUG~\cite{Sagonas2013}.
% Эти датасеты содежрат 2414 in-the-wild images with 3837 faces.
These datasets contain 2414 in-the-wild images with 3837 faces.
% Всего датасет 300W-LP содержит 122450 изображений.
\optionalTwo{The 300W-LP dataset}{300W-LP} contains 122450 images.
% Датасет AFLW содержит 21120 изображений, gathered from Flickr. 
\optionalTwo{The AFLW dataset}{AFLW} contains 21120 images gathered from Flickr.
% Датасет AFLW2000 содержит первые 2000 изображений из датасета AFLW.
\optionalTwo{The AFLW2000 dataset}{AFLW2000} contains the first 2000 images from the AFLW dataset.
% Датасет BIWI содержит 15677 изображений из 24 видео 20 персон, полученных в лаборатоных условиях с помощью Kinect.
\optionalTwo{The BIWI dataset}{BIWI} contains 15677 images from 24 videos of 20 people obtained in a laboratory using Kinect.
% Отметим, что датасет BIWI, в отличие от остальных, не содержит разметки bounding box.
Note that \optionalTwo{the BIWI dataset}{BIWI}, unlike the others, does not contain bounding boxes.
% Примеры изображений датасетов приведены на рисунке~\ref{fig_datasets}.
Examples of dataset images are shown in Fig.~\ref{fig_datasets}.

\subsection{Preprocessing} \label{ds_prep}

% Следуя ранее представленным работам \cite{Ruiz2018,Yang2019,Wang2019,Huang2020,Zhang2020} мы сократили датасеты, оставив только изображения с углами поворота головы в диапазоне $\left[-99^\circ; +99^\circ \right]$.
% Following~\cite{Ruiz2018,Yang2019,Wang2019,Huang2020,Zhang2020}, we reduced the datasets, leaving only images with head rotation angles in the range $\left[-99^\circ; +99^\circ \right]$.
We reduced the datasets following~\cite{Ruiz2018,Yang2019,Wang2019,Huang2020,Zhang2020}, leaving only images with head rotation angles in the range $\left[-99^\circ; +99^\circ \right]$.

% Изображения из датасетов содержат лицо и некоторый фон.
The images from the datasets contain a face and some background.
% Основным объектом анализа в задаче head pose estimation  является лицо.
The main object of analysis in the HPE problem is the face.
% Поэтому необходимо обратить внимание нейронной сети на лицо.
Therefore, it is necessary to draw the attention of the NN to it.
% Для этого производится обрезка изображения по bounding box, содержащем лицо и дальнейший ресайз до размеров нейронной сети.
The attention mechanism is performed by cropping the image on the bounding box containing the face and further resizing it to the size of the NN.

% В датасетах 300W-LP, AFLW2000 и AFLW разметка bounding box получена на основе построения ограничивающего прямоугольника вокруг разметки ключевых точек с последующим его небольшим увеличением.
The bounding boxes of 300W-LP\optional{~\cite{Zhu2016}}, AFLW2000\optional{~\cite{Zhu2016}}, and AFLW\optional{~\cite{Kostinger2011}} \optional{datasets} are obtained by constructing a bounding rectangle around the facial landmarks and then slightly enlarging them.
% Датасет BIWI не содержит разметку bounding box.
\optional{The BIWI dataset}{BIWI} does not contain bounding boxes.
% В работах по head pose estimation разметка получена разными способами.
In the literature on HPE, bounding boxes are obtained in different ways. 
% Так, в работе \cite{Ruiz2018} для этой задачи использовался детектор Faster R-CNN \cite{Ren2017}, в работе \cite{Yang2019} использовался детектор MTCNN \cite{Zhang2016}, в работах \cite{Zhang2020,Huang2020} использовался детектор Dlib \cite{King2009}, в работе \cite{Valle2020} использовались карты глубины.
For example, the Faster R-CNN~\cite{Ren2017} detector was used in~\cite{Ruiz2018}, the MTCNN~\cite{Zhang2016} detector was used in~\cite{Yang2019}, the Dlib~\cite{King2009} detector was used in~\cite{Zhang2020,Huang2020}, and the depth maps were used in~\cite{Valle2020}.

% Так как выбор bounding box влияет на результат \cite{Shao2019,Xue2020}, то различный способы получения bounding box не позволяет проводить корректную валидацию методов между собой. 
Since bounding boxes affect the result~\cite{Shao2019,Xue2020}, the differences in the ways of getting the bounding boxes do not allow for a valid comparison of the methods for HPE.
% Чтобы решить данную проблему, нами был обучен детектор YOLO v5 \cite{glenn_jocher_2021_4418161} на датасете WIDER \cite{Yang2016} и переразмечены с его помощью все четыре датасета: 300W-LP \cite{Zhu2016}, AFLW2000 \cite{Zhu2016}, AFLW \cite{Kostinger2011} и BIWI \cite{Fanelli2013}.
% To solve this problem, we trained the YOLOv5~\cite{glenn_jocher_2021_4418161} detector on the WIDER~\cite{Yang2016} dataset. 
% To solve this problem, we trained the YOLOv5~\cite{yolov5_source} detector on the WIDER~\cite{Yang2016} dataset. 
We trained the YOLOv5~\cite{yolov5_source} detector on the WIDER~\cite{Yang2016} dataset to solve this problem 
We redefined the bounding boxes using this detector for all four datasets: 300W-LP\optional{~\cite{Zhu2016}}, AFLW2000\optional{~\cite{Zhu2016}}, AFLW\optional{~\cite{Kostinger2011}}, and BIWI\optional{~\cite{Fanelli2013}}.
% Выполненная разметка датасетов выложена в открытом доступе \cite{angles_source}, что позволит выполнять более объективное сравнение алгоритмов решения задачи HPE.
The new bounding boxes for these datasets are publicly available in~\cite{angles_source}, which will allow performing an objective comparison of methods for the HPE problem without depending on the bounding boxes.

\subsection{Augmentations} \label{ds_aug}

% Неотъемлемой частью обучения нейронных сетей является аугментация данных. 
Data augmentation is an integral part of NN training.
% Применение аугментации позволяет существенно расширить обучающую выборку и повысить репрезентативность данных. 
Augmentation can significantly expand the training set and increase the representativeness of the data.
% За счёт использования аугментаций удаётся существенно повысить качество обученной нейронной сети \cite{Wang2017,Mikoajczyk2018}.
Due to augmentation, it is possible to significantly improve the quality of the trained NN~\cite{Wang2017,Mikoajczyk2018}.

% Применительно к head pose estimation problem аугментации можно разделить на два вида: влиящие и не влиящие на углы поворота головы.
Concerning the HPE problem, augmentation can be divided into two types: those that affect the head pose and those that do not affect the head pose.
% В нашем workflow сначала применяются аугментации влияющие на углы, а затем аугментации не влияющие на углы. 
In our workflow, augmentations that affect the head pose are applied first, and then augmentations that do not affect the head pose.

% Из аугментаций, влияющих на углы, мы используем отражение по горизонтали с вероятностью $0.5$ и вращение на случайный угол в диапазоне $\left[-15^{\circ}; +15^{\circ}\right]$. 
We use a horizontal flip with a probability of $0.5$ and rotate by a random angle in the range $\left[-15^{\circ}; +15^{\circ}\right]$ from the augmentations affecting the head pose.
% В процессе применения данных аугментаций происходит корректировка углов, описанная в работе \cite{Sheka2021}. 
In the process of applying these augmentations, the head poses are corrected, as described in~\cite{Sheka2021}. 

% \newpage
% Аугментации, не влияющие на углы, были взяты из библиотеки Albumentations \cite{Buslaev2020}.
The augmentations that do not affect the head pose were taken from the Albumentations library~\cite{Buslaev2020}.
% Сначала с вероятностью $0.5$ применяется одна из: сдвиг значения каждого HSV-канала на случайное число, соляризация, поканальная нормализация, contrast limited adaptive histogram equalization.
% First, one of the augmentations is applied with a probability of $0.5$: \texttt{HueSaturationValue(hue\_shift\_limit=20, sat\_shift\_limit=15, val\_shift\_limit=20)}, \texttt{CLAHE}, \texttt{Equalize(by\_channels=True)},~and \texttt{Solarize(threshold=240)}.
% First, one of the following augmentations is applied with a probability of $0.5$: \texttt{HueSaturationValue}, \texttt{CLAHE}, \texttt{Equalize},~and \texttt{Solarize}.
% Сначала с вероятностью 0.5 не применяется аугментация, либо с вероятностью 0.5 применяется одна случайно выбранная аугментация из списка: HueSaturationValue, CLAHE, Equalize, Solarize.
First, with a probability of $0.5$, no augmentation is applied, or with a probability of $0.5$, one randomly selected augmentation from the list is applied:  \texttt{HueSaturationValue}, \texttt{CLAHE}, \texttt{Equalize},~or \texttt{Solarize}.
% 
% 
% Затем с вероятностью $0.5$ применяется перемешивание или выброс 1-2 каналов. 
% Then either  \texttt{ChannelShuffle} or \texttt{ChannelDropout} are applied with a probability of $0.5$.
Then, with a probability of $0.5$, no augmentation is applied, or with a probability of $0.5$, one randomly selected augmentation from the list is applied: \texttt{ChannelShuffle} or \texttt{ChannelDropout}.
% Затем с вероятностью $0.5$ применяется размытие медианным фильтром.
% Then \texttt{MedianBlur(blur\_limit=3)} is applied with a probability of $0.5$.
Then \texttt{MedianBlur} is applied with a probability of $0.5$.
% Далее с вероятностью $0.5$ применяется добавление тени либо тумана либо цветного шума.
% Then either \texttt{RandomShadow} or \texttt{ISONoise} is applied with a probability of $0.5$.
Then, with a probability of $0.5$, no augmentation is applied, or with a probability of $0.5$, one randomly selected augmentation from the list is applied: \texttt{RandomShadow} or \texttt{ISONoise}.
% Наконец, с вероятностью $0.5$ применяется добавление 1 или 2 квадратов $16 \times 16$ в случайное место на изображении.
% Finally, \texttt{CoarseDropout(min\_holes=1, max\_holes=2, min\_width=16, max\_width=16, min\_height=16, max\_height=16)} is applied with a probability of $0.5$.
Lastly, \texttt{CoarseDropout} is applied with a probability of $0.5$.

\subsection{Testing protocols} \label{ds_proto}

% Согласно ранее представленным работам, мы используем следующие тестовые протоколы, приведённые в таблице~\ref{tab_protocols}. 
Following previous 
% {\color{red} papers} 
\optGreen{works}, we use the following test protocols, shown in Table~\ref{tab_protocols}.
% Ниже протоколы описаны более подробно.
The test protocols are described in more detail below.

\begin{table}[htbp]
    \centering
    % \caption{Тестовые протоколы. Оставлены изображения с углами поворота в диапазоне $\left[-99^\circ; +99^\circ \right]$}
    \caption{Test protocols}
    \label{tab_protocols}
    \begin{tabular}{c|cc|cc}
        \hline
        \multirow{2}{*}{Protocol} & \multicolumn{2}{c|}{Train} & \multicolumn{2}{c}{Test} \\
        \cline{2-5}
         & Dataset & \#  of Faces& Dataset & \# of Faces\\
        \hline
        \multirow{2}{*}{Protocol \#1} & \multirow{2}{*}{300W-LP} & \multirow{2}{*}{$122415$} & AFLW2000 & 1969 \\
         & & & BIWI & $15677$ \\
        % \hline
        Protocol \#2 & AFLW & $22079$ & AFLW & $1969$ \\
        % \hline
        Protocol \#3 & BIWI & $10612$ & BIWI & $5065$ \\
        \hline
    \end{tabular}

\end{table}

% \textbf{Тестовый протокол 1.}
\textbf{Test protocol 1.}
% Для обучения используется 300W-LP, для валидации используются AFLW2000 и BIWI.
300W-LP is used for training, AFLW2000 and BIWI are used for testing.
% Используются только те изображения датасетов, в которых каждый из углов поворота головы находится в диапазоне $\left[-99^\circ; +99^\circ \right]$.
Only images from datasets with head rotation angles in the range $\left[-99^\circ; +99^\circ \right]$ were used.

% \textbf{Тестовый протокол 2.}
\textbf{Test protocol 2.}
% Датасет AFLW используется для обучения и валидации.
\optionalTwo{The AFLW dataset}{AFLW} is used for training and testing.
% В работе \cite{Amador2018} указано, что для датасета AFLW нет стандартного протокола.
In~\cite{Amador2018}, 
% {\color{red}it was pointed out} 
\optGreen{\citeauthor{Amador2018} point out}  
no standard test protocol for \optionalTwo{the AFLW dataset}{AFLW}.
% Кроме того, в некоторых работах \cite{Kumar2018,Wang2019} используют случайное разделение датасета на trainset и testset, что не позволяет проводить корректное сравнение.
In addition, some 
% {\color{red}papers} 
\optGreen{researchers}~\cite{Kumar2018,Wang2019} use a random division of the dataset into training and testing.
It does not allow for a valid comparison of methods.
% По этим причинам мы решили для валидации использовать первые 2000 jpg-изображений, составляющие датасет AFLW2000, остальные использовать для обучения.
For these reasons, we decided to use the first 2000 jpg images that make up \optionalTwo{the AFLW2000 dataset}{AFLW2000} for testing and use the rest for training.
% Предварительно были оставлены только изображения датасетов с углами поворота в диапазоне $\left[-99^\circ; +99^\circ \right]$.
Similarly, only images from datasets with head rotation angles in the range $\left[-99^\circ; +99^\circ \right]$ were used.

% \textbf{Тестовый протокол 3.}
\textbf{Test protocol 3.}
% Для обучения используются изображения, соответствующие 70\% видео из датасета BIWI.
Images corresponding to 70\% of the video from \optionalTwo{the BIWI dataset}{BIWI} are used for training.
% Остальные изображения используются для валидации.
The remaining images are used for testing.
% Разделение датасета на trainset и testset выполнено согласно работе \cite{Yang2019}.
The division of the dataset into training and testing is performed following the method of~\cite{Yang2019}.
% Аналогично, используются только изображения датасета с углами поворота в диапазоне $\left[-99^\circ; +99^\circ \right]$.
Similarly, only images from datasets with head rotation angles in the range $\left[-99^\circ; +99^\circ \right]$ were used.

\section{Experiments}
% В данном разделе мы опишем детали экспериментов. 
In this section, we describe the experiments.
% В подразделе описано окружение в котором проводились эксперименты. 
Subsection~\ref{impl_details} describes the environment in which the experiments were conducted.
% Подраздел содержит результаты экспериментов для 1й стадии.
Subsection~\ref{stage1_exp} contains the results of the experiments for the first stage.
% Подраздел содержит результаты экспериментов для 2й стадии.
Subsection~\ref{stage2_exp} contains the results of the experiments for the second stage.

\subsection{Implementation details}  \label{impl_details}

% Python\optional{~\cite{van2007python}} was used to implement these computational experiments.
% The NNs were implemented in the Pytorch\optional{~\cite{Ketkar2017}} library.
% The data augmentation was performed using the Albumentations library.
% Our research was performed using ``Uran'' supercomputer of the IMM UB RAS. 
% The NNs were trained on 8 Tesla v100 GPUs with 32GB of memory.

We implemented NNs using Pytorch and Albumentations libraries.
Our research was performed using the ``Uran'' supercomputer of the IMM UB RAS. 
The NNs were trained on 8 Tesla v100 GPUs with 32GB of memory.

\subsection{First stage}  \label{stage1_exp}

% На первом этапе при обучении нейронной сети в качестве backbone использовался Resnet50, предобученный на ImageNet \cite{JiaDeng2009}. 
In the first stage, we used ResNet50 as a backbone, pre-trained on ImageNet~\cite{JiaDeng2009}. 
% Нейронная сеть содержала 5 head branches: 4 классификации с размерами бинов $B = \left\{1,2,3,4\right\}$ и регрессию.
The NN contained 5 head branches: 4 classification heads with bin sizes $B = \left\{1,2,3,4\right\}$ and a regression head.

% The NN was trained using the Adam~\cite{Kingma2015} optimizer with a learning rate of $10^{-4}$ for 100 epochs.
% The batch size was 128.
The NN was trained using the Adam~\cite{Kingma2015} optimizer with a batch size of 128 and a learning rate of $10^{-4}$ for 100 epochs.

% {
% \color{blue}
% % Для построения ансамбля были взяты параметры $K$ и $d$ из диапазона $\left[0;15\right]$.
% To construct the offset ensemble, the parameters $K$ and $d$ were taken from the range $\left[0;15\right]$.
% % Отметим, что при $K < d$ размер ансамбля равен 1, т.е. фактически он не используется.
% Note that for $K < d$, the size of the ensemble is 1, i.e. it is not actually used.
% }

\optGreen{To construct the convolutional ensemble, the parameters $s \in \left[0;15\right]$ and $p \in \left\{0, s, 2s, ...\right\}, p \leq 15$ were taken.
Note that for $p = 0$, the size of the ensemble is 1, i.e. it is not actually used.
% Note that for $p = 0$, the size of the ensemble is 1, i.e. it is not used.
}

\begin{figure}[ht!bp]
% \begin{figure*}[htb!]
% \begin{figure*}[ht!]
    \centering
    \begin{subfigure}[b]{\linewidth}
        \centering
        \includegraphics[width=\textwidth]{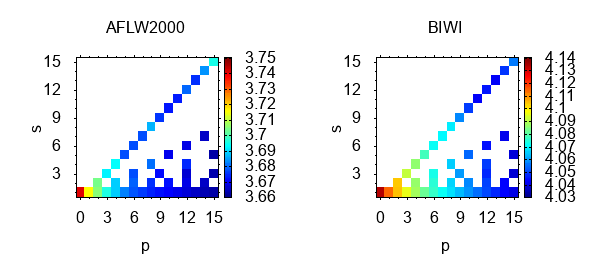}
        \subcaption*{Protocol 1}
    \end{subfigure}
    \hfill
    \begin{subfigure}[b]{.49\linewidth}
        \centering
        \includegraphics[width=\textwidth]{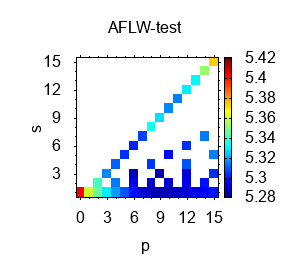}
        \subcaption*{Protocol 2}
    \end{subfigure}
    \hfill
    \begin{subfigure}[b]{.49\linewidth}
        \centering
        \includegraphics[width=\textwidth]{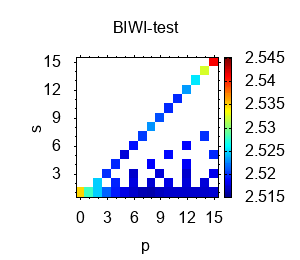}
        \subcaption*{Protocol 3}
    \end{subfigure}
    \caption{
    MAE heat map of results \optGreen{for various parameters of the convolutional ensemble: 
    $p$ defines the size of the receptive field, $s$ defines the sparsity.
    Increasing $s$ decreases computations, and increasing $p$ increases computations.
    The figure contains white spots because $s$ must be a divisor of $p$.
    }
    }
    \label{fig_ens_aps}
\end{figure}

% В таблицах~\ref{tab_ens_ap1},~\ref{tab_ens_ap2},~\ref{tab_ens_ap3} для каждого тестового протокола приведены соответствующие численные значения метрики MAE для некоторых параметров $K$ и $d$.
Tables~\ref{tab_ens_ap1},~\ref{tab_ens_ap2}, and~\ref{tab_ens_ap3} for each test protocol show the corresponding values of the MAE metric for some parameters 
% {\color{red}$K$ and $d$} 
\optGreen{$s$ and $p$}.
$A$ is the size of the ensemble.
% Значение метрики MAE при различных значениях параметров $K$, $d$ для всех трёх тестовых протоколов показано в виде тепловых карт на рисунке~\ref{fig_ens_aps}.
Fig.~\ref{fig_ens_aps} shows the MAE heat map for different values of the parameters 
% {\color{red}$K$, $d$} 
\optGreen{$s$, $p$} for all three test protocols.
% Жирным выделены наилучшие результаты.
The best results are highlighted in bold.
% Наилучшие параметры ансамбля также выделены жирным.
The best parameters of the ensemble are also highlighted in bold.

\begin{table}[ht!]
    \centering
    % \caption{Результаты MAE ансамбля для тестового протокола 1}
    \caption{MAE ensemble results for test protocol 1}
    \label{tab_ens_ap1}
    \setlength{\tabcolsep}{1.7pt}
    \begin{tabular}{cc|c|ccc|c|ccc|c}
        \hline
        \multirow{2}{*}{\optGreen{$p$}} & \multirow{2}{*}{\optGreen{$s$}} & \multirow{2}{*}{$A$} &
        \multicolumn{4}{c|}{AFLW2000} & \multicolumn{4}{c}{BIWI} \\
        \cline{4-11}
         & & & pitch & yaw & roll & avg. & pitch & yaw & roll & avg. \\
        \hline
        $0$\rlap{$^{\mathrm{a}}$} & $1$ & $1$ & $4.861$ & $3.064$ & $3.300$ & $3.742$ & $5.181$ & $4.119$ & $3.105$ & $4.135$ \\
        $8$ & $1$ & $289$ & $\bf{4.813}$ & $2.980$ & $3.231$ & $3.674$ & $\bf{5.171}$ & $4.004$ & $3.029$ & $4.068$ \\
        $9$ & $3$ & $49$ & $4.814$ & $2.974$ & $3.227$ & $3.671$ & $5.182$ & $3.978$ & $3.022$ & $4.061$ \\
        $15$ & $1$ & $961$ & $4.819$ & $\bf{2.954}$ & $3.218$ & $\bf{3.664}$ & $5.211$ & $3.907$ & $3.006$ & $4.041$ \\
        $\bf{15}$ & $\bf{5}$ & $49$ & $4.817$ & $2.959$ & $\bf{3.216}$ & $\bf{3.664}$ & $5.226$ & $3.888$ & $\bf{3.005}$ & $\bf{4.040}$ \\
        $15$ & $15$ & $9$ & $4.841$ & $2.998$ & $3.248$ & $3.695$ & $5.265$ & $\bf{3.877}$ & $3.029$ & $4.057$ \\
        \hline
        % \multicolumn{10}{l}{$^{\mathrm{a}}${baseline, ансамбль не используется}} \\
        \multicolumn{11}{l}{$^{\mathrm{a}}${baseline, the ensemble is not used}} \\
    \end{tabular}

\end{table}

\begin{table}[ht!]
    \centering
    % \caption{Результаты MAE ансамбля для тестового протокола 2}
    \caption{MAE ensemble results for test protocol 2}
    \label{tab_ens_ap2}
    \begin{tabular}{cc|c|ccc|c}
        \hline
        \optGreen{$p$} & \optGreen{$s$} & $A$ & pitch & yaw & roll & avg. \\
        \hline
        $0$\rlap{$^{\mathrm{a}}$} & $1$ & $1$ & $6.419$ & $5.375$ & $4.416$ & $5.404$ \\
        $8$ & $1$ & $289$ & $\bf{6.268}$ & $5.289$ & $4.307$ & $5.288$ \\
        $\bf{9}$ & $\bf{3}$ & $49$ & $6.274$ & $5.290$ & $\bf{4.296}$ & $\bf{5.286}$ \\
        $15$ & $1$ & $961$ & $6.310$ & $\bf{5.287}$ & $4.304$ & $5.300$ \\
        $15$ & $5$ & $49$ & $6.341$ & $5.290$ & $4.315$ & $5.315$ \\
        $15$ & $15$ & $9$ & $6.412$ & $5.355$ & $4.357$ & $5.375$ \\
        \hline
        % \multicolumn{6}{l}{$^{\mathrm{a}}${baseline, ансамбль не используется}} \\
        \multicolumn{7}{l}{$^{\mathrm{a}}${baseline, the ensemble is not used}} \\
    \end{tabular}
\end{table}

\begin{table}[ht!]
    \centering
    % \caption{Результаты MAE ансамбля для тестового протокола 3}
    \caption{MAE ensemble results for test protocol 3}
    \label{tab_ens_ap3}
    \begin{tabular}{cc|c|ccc|c}
        \hline
        \optGreen{$p$} & \optGreen{$s$} & $A$ & pitch & yaw & roll & avg. \\
        \hline
        $0$\rlap{$^{\mathrm{a}}$} & $1$ & $1$ & $2.862$ & $2.593$ & $2.150$ & $2.535$ \\
        $\bf{8}$ & $\bf{1}$ & $289$ & $\bf{2.840}$ & $2.569$ & $\bf{2.141}$ & $\bf{2.516}$ \\
        $9$ & $3$ & $49$ & $\bf{2.840}$ & $2.569$ & $2.144$ & $2.518$ \\
        $15$ & $1$ & $961$ & $2.841$ & $2.559$ & $2.153$ & $2.518$ \\
        $15$ & $5$ & $49$ & $2.847$ & $\bf{2.556}$ & $2.157$ & $2.520$ \\
        $15$ & $15$ & $9$ & $2.876$ & $2.566$ & $2.183$ & $2.542$ \\
        \hline
        % \multicolumn{6}{l}{$^{\mathrm{a}}${baseline, ансамбль не используется}} \\
        \multicolumn{7}{l}{$^{\mathrm{a}}${baseline, the ensemble is not used}} \\
    \end{tabular}

\end{table}

% Отметим, что исходя из формулы~\eqref{bboxes} размер ансамбля квадратично зависит от значения $K$ и обратно пропорционален значению $d$.
The size of the ensemble depends quadratically on the value of 
% {\color{red}$K$}
\optGreen{$p$} and is inversely proportional to the square of %the value of 
% {\color{red}$d$}
\optGreen{$s$}.
% , in accordance with~\eqref{bboxes}
% Большой размер ансамбля может существенно снизить производительность.
Having a large ensemble can significantly reduce the performance.
% Данный факт необходимо учитывать при выборе оптимальных значений $K$ и $d$.
Therefore, it is necessary to consider the size of the ensemble when choosing the optimal values of
% {\color{red}$K$ and $d$}
\optGreen{$s$ and $p$}.

\subsection{Second stage} \label{stage2_exp}

% Сначала на основе результатов первого этапа были определены наилучшие параметры ансамбля.
Based on the results of the first stage, the best parameters of the convolutional ensemble were determined.
% При значениях $K=15$ и $d=5$ достигается максимальное улучшение на первом тестовом протоколе.
The maximum improvement is achieved on the first test protocol with  
% {\color{red}$K=15$ and $d=5$}
\optGreen{$p=15$ and $s=5$}.
% Другие протоколы показывают при данных параметрах результат, сравнимый с оптимальным.
Other protocols have results comparable to the optimal one with these parameters.

% При обучении на втором этапе в качестве backbone использовались различные архитектуры семейства Resnet: Resnet18, Resnet34, Resnet50, Resnet101 и Resnet152.
In the second stage, various architectures of the ResNet family were used as the backbone: ResNet18, ResNet34, ResNet50, ResNet101 and ResNet152.
% Все backbone предобученны на ImageNet.
All backbones were pre-trained on ImageNet\optional{~\cite{JiaDeng2009}}.
% Также использовался один head branch: классификация с размером бина 1.
A single head branch was also used: a classification head with a bin size of 1.

The NN was trained using the Adam\optional{~\cite{Kingma2015}} optimizer with a learning rate of $3 \cdot 10^{-4}$ for 100 epochs.
The batch size was 128.

% В таблице \ref{comp_sota} приведены сводные результаты экспериментов для каждого тестового протокола и их сравнение со SOTA результатами.
Table~\ref{comp_sota} summarizes the results of the experiments for each test protocol and their comparison with the \optionalTwo{state-of-the-art}{SOTA} results.
% Наилучшие результаты в таблице выделены жирным.
The best results in Table~\ref{comp_sota} are highlighted in bold.

\begin{figure}[ht]
    \centering
    \includegraphics[width=\linewidth]{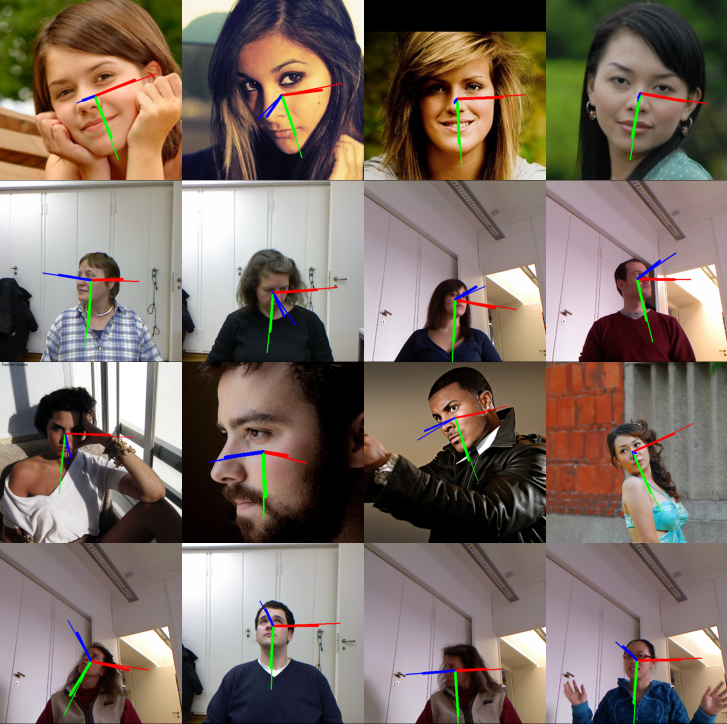}
    % \caption{Визуализация предсказаний для ResNet152. \\
    % Первые две строки соответствуют протоколу 1, датасеты AFLW2000 и BIWI соответственно. \\
    % Вторая строка соответствует протоколу 2, датасет AFLW. \\
    % Третья строка соответствует протоколу 3, датасет BIWI. \\
    % Жирные линии соответствуют предсказанию, тонкие соответствуют ground truth.}
    \caption{Visualization of ResNet152 predictions. 
    The first two lines correspond to Test protocol 1, AFLW2000, and BIWI datasets, respectively. 
    The second line corresponds to Test protocol 2, the AFLW dataset. 
    The third line corresponds to Test protocol 3, the BIWI dataset. 
    Bold lines correspond to the prediction, and thin lines correspond to the ground truth.}
    \label{fig_pred}
\end{figure}

% На рисунке~\ref{fig_pred} приведены примеры предсказаний обученной нейронной сети с backbone Resnet152 для всех тестовых протоколов.
Fig.~\ref{fig_pred} shows examples of predictions of a trained NN with ResNet152 backbone for all test protocols.

\section{Discussion}

% В данном разделе обсудим полученные результаты для каждого тестового протокола и каждого элемента of the KD workflow.
% \optCyan{In this section, we discuss the results obtained for each test protocol and each element of the KD workflow.}
\optCyan{This section discusses the results obtained for each test protocol and each element of the KD workflow.}

\subsection{Test protocol 1}

% Для первого тестового протокола использование ансамбля даёт прирост качества предсказания с $3.74$ до $3.67$ ($1.9\%$) для датасета AFLW2000 и с $4.14$ до $4.04$ ($2.4\%$) для датасета BIWI.
% In the first test protocol, the use of the convolutional ensemble gives an increase in the quality of the prediction from $ 3.74$ to $3.67$ ($1.9\%$) for \optionalTwo{the AFLW2000 dataset}{AFLW2000} and from $4.14$ to $4.04$ ($2.4\%$) for \optionalTwo{the BIWI dataset}{BIWI}.
In the first test protocol, the convolutional ensemble increases the quality of the prediction from $ 3.74$ to $3.67$ ($1.9\%$) for AFLW2000 and from $4.14$ to $4.04$ ($2.4\%$) for BIWI.
% Использование дистиляции позволяет дополнительно улучшить результаты.
The KD further improves the results.
% Улучшение достигается на всех вариантах backbone.
The improvement is achieved on all variants of the backbone.
% Для датасета AFLW2000 прослеживается улучшение результатов с ростом размера backbone.
For \optionalTwo{the AFLW2000 dataset}{AFLW2000}, there is an improvement in the results with an increase in the depth of the backbone. 
% Наилучший результат достигается для ResNet152.
The best result is achieved with the ResNet152 backbone.
% Использование дистиляции позволило увеличить качество предсказания по сравнению с вариантом без использования ансамбля с $3.74$ до $3.48$ ($7.0\%$) для датасета AFLW2000 и с $4.14$ до $3.70$ ($10.6\%$) для датасета BIWI.
The KD increases the prediction quality compared to the non-ensemble version from $ 3.74$ to $3.48$ ($7.0\%$) for \optionalTwo{the AFLW2000 dataset}{AFLW2000} and from $4.14$ to $3.70$ ($10.6\%$) for \optionalTwo{the BIWI dataset}{BIWI}.

% По сравнению со SOTA результатами наилучший показанный результат превосходит SOTA для датасета AFLW2000 с $3.83$ до $3.48$ ($9.1\%$).
The best result shown is superior to the \optionalTwo{state-of-the-art}{SOTA} results for \optionalTwo{the AFLW2000 dataset}{AFLW2000}. 
In particular, the increase is $9.1\%$: $3.48$ versus $3.83$.
% Для датасета BIWI получен результат, сравнимый со SOTA: разница составляет $1.0\%$.
The result comparable to the \optionalTwo{state-of-the-art}{SOTA} result is obtained for \optionalTwo{the BIWI dataset}{BIWI}: $3.70$ versus $3.66$, respectively. 
The difference is $1.0\%$.

% Результат, полученный на датасете BIWI можно объяснить тем, что предыдущий state-of-the-art результат был получен с неявным использованием дополнительных данных о ключевых точках. 
The result obtained on \optionalTwo{the BIWI dataset}{BIWI} can be explained by the previous \optionalTwo{state-of-the-art}{SOTA} result obtained with the implicit use of additional data about facial landmarks.
% У нас проводились чистые эксперименты согласно тестовому протоколу без использования дополнительных данных.
We conducted pure experiments according to the test protocol without using additional data.
% Также в нашей работе используется Yolo v5 детектор, который нормализует положения bounding boxes относительно лица во всех датасетах.
We also used the YOLOv5 detector, which normalizes the positions of the bounding boxes relative to the face in all datasets.
% Некоторые работы не содержат четкого описания получения bounding boxes для датасетов.
% Some publications do not contain a clear description of how the bounding boxes were obtained for the datasets. 
Some publications do not clearly describe how the bounding boxes were obtained for the datasets. 
% Как мы упомянали ранее, выбор bounding boxes может существенно улучшить результат без изменения метода ~\cite{Shao2019,Xue2020}. 
As we mentioned earlier, choosing the bounding boxes can significantly improve the result without changing the method~\cite{Shao2019,Xue2020}.

\subsection{Test protocol 2}

% Для второго тестового протокола отметим, что в разных работах используют разное разделение датасета AFLW на trainset и testset.
Different 
% {\color{red}papers} 
\optGreen{researchers} 
use a different division of \optionalTwo{the AFLW dataset}{AFLW} into training and testing parts for the second test protocol.
% Это опять же не позволяет объективно сравнивать между собой результаты алгоритмов.
It does not allow us to compare the results of different methods objectively.

% Использование ансамбля для данного протокола даёт прирост качества предсказания с $5.40$ до $5.32$ ($1.5\%$).
The convolutional ensemble for this protocol increases the prediction quality from $5.40$ to $5.32$ ($1.5\%$).
% Использование дистиляции позволяет увеличить качество предсказания по сравнению с вариантом без использования ансамбля с $5.40$ до $5.10$ ($5.6\%$).
The KD allows one to increase the prediction quality compared to the non-ensemble variant from $5.40$ to $5.10$ ($5.6\%$).
% Также отметим, что дистиляция улучшает варианты на всех рассмотренных вариантах backbone.
Also, note that KD improves the quality of the convolutional ensemble on all the backbone options considered.

\subsection{Test protocol 3}

% Для третьего тестового протокола использование ансамбля даёт несущественный прирост качества предсказания: оно составляет лишь $0.8\%$.
The convolutional ensemble for the third protocol slightly increases the prediction quality from $2.54$ to $2.52$ ($0.8\%$).
% Дистилиция знаний ухудшает точность по сравнению с использованием и без использования анасабмля. 
The KD decreases the prediction quality compared to ensemble and non-ensemble variants.
% Наилучший показанный результат превосходит state-of-the-art результат  с $2.80$ до $2.52$ ($10\%$).
The best result increases the \optionalTwo{state-of-the-art}{SOTA} result from  $2.80$ to $2.52$ ($10\%$).

% Отметим, что датасет BIWI имеет хорошую репрезентативность по head poses, но плохую репрезентативность по персонам и условиям съемки, так как он получен в лабораторных условиях и содержит лишь 20 персон. 
% Note that \optionalTwo{the BIWI dataset}{BIWI} has a high representativeness in terms of head poses but a low representativeness in terms of shooting conditions and people, since it was obtained in the laboratory, and it contains only 20 people.
Note that BIWI has high representativeness in terms of head poses but low representativeness in terms of shooting conditions and people since it was obtained in the laboratory and contains only 20 people.
% Видимо низкая репрезентативность по персонам и условиям съемки не позволяет проявить сильный стороны ансамбля смещений и дистиляции знаний.
Apparently, the low representativeness in terms of shooting conditions and people does not show the strengths of the convolutional ensemble and the KD.

\begin{table*}[ht!]
    \centering
    \caption{Comparison of head pose estimations}
    \label{comp_sota}
    \begin{tabularx}{\linewidth}{c|c||XXX|X|XXX|X||XXX|X||XXX|X}
        \hline
        \multicolumn{1}{c|}{\multirow{3}{*}{Method}} & Use & \multicolumn{8}{c||}{Protocol 1} & \multicolumn{4}{c||}{\multirow{2}{*}{Protocol 2}} & \multicolumn{4}{c}{\multirow{2}{*}{Protocol 3}} \\
        \cline{3-10}
        \multicolumn{1}{c|}{} & extra & \multicolumn{4}{c|}{AFLW2000} & \multicolumn{4}{c||}{BIWI} & \multicolumn{4}{c||}{} & \multicolumn{4}{c}{} \\
        \cline{3-18}
        & data & pitch & yaw & roll & avg. & pitch & yaw & roll & avg. & pitch & yaw & roll & avg. & pitch & yaw & roll & avg. \\
        \hline
        HopeNet \cite{Ruiz2018} & & $6.55$ & $6.47$ & $5.43$ & $6.15$ & $6.60$ & $4.81$ & $3.26$ & $4.89$ & $5.89$ & $6.26$ & $3.82$ & $5.32$ & $3.39$ & $3.29$ & $3.00$ & $3.23$ \\
        FSA-Net \cite{Yang2019} & & $6.08$ & $4.50$ & $4.64$ & $5.07$ & $4.96$ & $4.27$ & $2.76$ & $4.00$ & & & & & $4.29$ & $2.89$ & $3.60$ & $3.60$ \\
        QuatNet \cite{Hsu2019} & & $5.61$ & $3.97$ & $3.92$ & $4.50$ & $5.49$ & $4.01$ & $2.93$ & $4.14$ & $4.31$ & $\bf{3.93}$ & $2.59$ & $3.61$ & & & & \\
        Hybrid \cite{Wang2019} & & $6.28$ & $4.82$ & $5.14$ & $5.40$ &  &  &  &  &  &  &  & & & & & \\
        BBox Margin \cite{Shao2019} & & $6.37$ & $5.07$ & $4.99$ & $5.48$ & $7.25$ & $4.59$ & $6.15$ & $6.00$ &  &  &  & & & & & \\
        CNN + Heatmaps \cite{Gupta2019} & \checkmark & & & & & & & & & $4.43$ & $5.22$ & $2.53$ & $4.06$ & & & & \\
        FDN \cite{Zhang2020} & & $5.61$ & $3.78$ & $3.88$ & $4.42$ & $4.70$ & $4.52$ & $2.56$ & $3.93$ & & & & & $3.98$ & $3.00$ & $2.88$ & $3.29$ \\
        MNN \cite{Valle2020} & \checkmark  & $4.69$ & $3.34$ & $3.48$ & $3.83$ & $4.61$ & $3.98$ & $\bf{2.39}$ & $\bf{3.66}$ & $\bf{3.07}$ & $4.16$ & $\bf{2.43}$ & $\bf{3.22}$ & & & & \\
        % WHENet \cite{Zhou2020} & $5.75$ & $4.44$ & $4.31$ & $4.83$ & $4.10$ & $3.60$ & $2.73$ & $\bf{3.48}$ &  &  &  & & & & & \\
        HPE-40 \cite{Huang2020} & & $6.18$ & $4.87$ & $4.80$ & $5.28$ & $5.18$ & $4.57$ & $3.12$ & $4.29$ &  &  &  & & & & & \\
        TriNet \cite{Cao2020} & & $5.77$ & $4.20$ & $4.04$ & $4.67$ & $4.76$ & $\bf{3.05}$ & $4.11$ & $3.97$ & & & & & $3.04$ & $\bf{2.44}$ & $2.93$ & $2.80$ \\
        WSM-LgR \cite{Abate2020} & \checkmark & & & & & & & & & & & & & $4.56$ & $2.47$ & $\bf{2.13}$ & $3.05$ \\
        % Classification \cite{} & $4.65$ & $3.79$ & $3.13$ & $3.85$ &  &  &  &  &  &  &  & & & & & \\
        \hline
        Base ResNet18 & & $5.05$ & $3.07$ & $3.52$ & $3.88$ & $5.51$ & $4.52$ & $3.16$ & $4.40$ & $6.29$ & $5.37$ & $4.65$ & $5.44$ & $3.61$ & $2.85$ & $2.77$ & $3.08$ \\
        Base ResNet34 & & $4.95$ & $3.04$ & $3.38$ & $3.79$ & $5.19$ & $4.47$ & $3.35$ & $4.34$ & $6.21$ & $5.58$ & $4.66$ & $5.48$ & $3.07$ & $2.67$ & $2.49$ & $2.74$ \\
        Base ResNet50 & & $4.86$ & $3.06$ & $3.30$ & $3.74$ & $5.18$ & $4.12$ & $3.11$ & $4.14$ & $6.42$ & $5.38$ & $4.42$ & $5.40$ & $2.86$ & $2.59$ & $2.15$ & $2.54$ \\
        Base ResNet101 & & $5.03$ & $3.19$ & $3.42$ & $3.88$ & $5.38$ & $4.34$ & $3.17$ & $4.30$ & $6.17$ & $5.38$ & $4.40$ & $5.32$ & $2.56$ & $2.55$ & $2.34$ & $2.49$ \\
        Base ResNet152 & & $4.86$ & $3.13$ & $3.35$ & $3.78$ & $5.05$ & $4.09$ & $3.10$ & $4.08$ & $5.92$ & $5.54$ & $4.29$ & $5.25$ & $2.71$ & $2.55$ & $2.19$ & $2.48$ \\
        \hline
        % Ensemble: $\left(K, d\right)=\left(15, 5\right)$ & $4.82$ & $2.96$ & $3.22$ & $3.67$ & $5.23$ & $3.89$ & $3.01$ & $4.04$ & $6.34$ & $5.29$ & $4.32$ & $5.32$ & $2.85$ & $2.56$ & $2.16$ & $\bf{2.52}$ \\
        Ensemble ResNet18 & & $4.94$ & $3.00$ & $3.40$ & $3.78$ & $5.48$ & $4.40$ & $3.09$ & $4.32$ & $6.25$ & $5.32$ & $4.49$ & $5.35$ & $3.62$ & $2.85$ & $2.76$ & $3.08$ \\
        Ensemble ResNet34 & & $4.95$ & $2.98$ & $3.34$ & $3.76$ & $5.19$ & $4.39$ & $3.31$ & $4.30$ & $6.12$ & $5.51$ & $4.58$ & $5.40$ & $3.04$ & $2.65$ & $2.49$ & $2.73$ \\
        Ensemble ResNet50 & & $4.82$ & $2.96$ & $3.22$ & $3.67$ & $5.23$ & $3.89$ & $3.01$ & $4.04$ & $6.34$ & $5.29$ & $4.32$ & $5.32$ & $2.85$ & $2.56$ & $2.16$ & $2.52$ \\
        Ensemble ResNet101 & & $4.99$ & $3.12$ & $3.42$ & $3.85$ & $5.31$ & $4.25$ & $3.11$ & $4.22$ & $6.09$ & $5.30$ & $4.29$ & $5.23$ & $\bf{2.53}$ & $2.54$ & $2.31$ & $\bf{2.46}$ \\
        Ensemble ResNet152 & & $4.85$ & $3.06$ & $3.32$ & $3.74$ & $5.01$ & $3.98$ & $3.05$ & $4.01$ & $5.77$ & $5.44$ & $4.21$ & $5.14$ & $2.70$ & $2.56$ & $2.17$ & $2.48$ \\
        \hline
        KD-ResNet18 & & $4.69$ & $3.00$ & $3.22$ & $3.64$ & $5.07$ & $3.96$ & $3.06$ & $4.03$ & $6.02$ & $5.45$ & $4.16$ & $5.21$ & $2.82$ & $2.59$ & $2.34$ & $2.58$ \\
        KD-ResNet34 & & $4.74$ & $3.00$ & $3.21$ & $3.65$ & $\bf{4.49}$ & $4.06$ & $2.96$ & $3.83$ & $6.10$ & $5.39$ & $4.24$ & $5.24$ & $2.93$ & $2.59$ & $2.35$ & $2.62$ \\
        KD-ResNet50 & & $4.68$ & $\bf{2.92}$ & $3.11$ & $3.57$ & $4.80$ & $4.05$ & $2.94$ & $3.93$ & $6.04$ & $5.38$ & $4.28$ & $5.24$ & $2.87$ & $2.73$ & $2.34$ & $2.65$ \\
        KD-ResNet101 & & $4.65$ & $2.96$ & $3.02$ & $3.54$ & $5.31$ & $3.76$ & $2.76$ & $3.94$ & $5.96$ & $5.36$ & $3.98$ & $5.10$ & $2.76$ & $2.74$ & $2.25$ & $2.58$ \\
        KD-ResNet152 & & $\bf{4.52}$ & $2.97$ & $\bf{2.96}$ & $\bf{3.48}$ & $4.73$ & $3.50$ & $2.87$ & $3.70$ & $5.93$ & $5.41$ & $4.07$ & $5.14$ & $2.88$ & $2.61$ & $2.37$ & $2.62$ \\
        \hline
        % \multicolumn{9}{l}{$^{\mathrm{*}}${Additional data with facial landmarks have been used during the training.}} \\
    \end{tabularx}
\end{table*}

\optCyan{\subsection{Neural network and data preprocessing}

% Предобработка данных является важной частью обучения нейронной сети.
Data preprocessing is an important part of NN training.
% Ранее мы показали, что использование аугментации вращения и нейронной сети с одной RvC головой позволяет получить результат, который незначительно уступает только MNN. 
Earlier, we demonstrated~\cite{Sheka2021} that using rotation augmentation and a NN with one RvC head allows us to get a result that is slightly inferior only to MNN.
% В частности, для первого тестового протокола на датасете AFLW2000 была получена ошибка  равная 3.85 MAE.
% В частности, MAE равна 3.85 в первом тестовом протоколе на датасете AFLW2000.
In particular, MAE is equal to 3.85 in the first test protocol on \optionalTwo{the AFLW2000 dataset}{AFLW2000}.
% Без использования аугментации вращения ошибка составила 4.95, что лучше многих предыдущих работ~\cite{}. 
% MAE равна 4.95 без использования аугментации вращения, что лучше многих предыдущих работ~\cite{}. 
MAE is equal to 4.95 without rotation augmentation, which is better than many previous works~\cite{Ruiz2018,Yang2019,Wang2019,\optional{Shao2019,}Huang2020}.
% Данный прирост был получен за счёт использование разметки прямоугольников, полученные с помощью детектора YOLOv5.
This accuracy is increased by bounding boxes relabeling obtained by the YOLOv5 detector.

% В рамках данной работы мы добавили дополнительные аугментации и доработали архитектуру нейронной сети.
In this paper, we have added additional augmentations and improved the NN architecture.
% Дополнительные аугментации манипулируют с цветовой схемой и добавляют шум.
Additional augmentations manipulate the color scheme and add noise.
% В нейронную сеть мы добавили дополнительные головы, которые выполнянли роль регулязаторов при обучении.
We have added additional heads, which provide regularization during the NN training.
% При этом, удаление любой головы увеличивало ошибку.
At the same time, removing any head branch increased the error.
% Добавленные улучшения уменьшили ошибку для архитектуры ResNet50 до 3.74 на датасете AFLW2000, что лучше любой предыдущей работы.
The improvements have reduced MAE for ResNet50 to 3.74 on AFLW2000.
It is better than any previous job with the same backbone.

% ResNet50 показывает результаты хуже некоторых работ~\cite{} на датасете BIWI в 1ом тестовом протоколе.
Base ResNet50 results worse than some works~\cite{Yang2019,Zhang2020,Valle2020,Cao2020} on BIWI in the 1st test protocol.
% Это обусловлено низкой репрезентативности 300w-lp отностительно BIWI. 
It is due to the low representativeness of 300W-LP relative to BIWI.
% В 3ем тестовом протоколе, полностью состоящего из BIWI, предложенная архитектура показывает результаты лучше всех предыдущих. 
In the 3rd test protocol, based on BIWI, the proposed architecture achieves better results than all the previous ones.
% ВПредложенная схема кодирования углов менее устойчива к выбросам, но лучше подходит для дистилляции знаний.
The proposed angle coding scheme is less resistant to outliers but is better suited for KD.

\subsection{Convolutional ensemble}

% Предложенный ансамбль повышает точность предсказания, использую одну обученную нейронную сеть.
The proposed ensemble increases prediction accuracy using a single trained NN.
% Ансамбль предсказывает значения с помощью нейронной сети по регулярной сетке как свёртка.
% {\color{Red}The ensemble predicts using NN over a regular grid as a convolution.}
% Формирование широкой нейронной сети не позволяет добиться нужно результата.
% A wide alternative NN does not improve the result.
A wide alternative NN operating with an enlarged image does not improve the result.
% The NN does not receive critical information about the image borders.
As described in section~\ref{method_ensemble}, the NN does not receive critical information about the image borders by padding function.
% В некотором смысле, свёрточный ансамбль является разновидностью test time augmentation.
In a sense, the convolutional ensemble is a kind of test time augmentation.

% На всех тестовых протоколах сверточный ансамбль увеличивает точность, который естественным образом сглаживает выбросы.
The convolutional ensemble increases accuracy on all test protocols by smoothing off outliers.
% Наибольший прирост ансамбль имеет на 1ом тестом протоколе, имеющий большой обучающий датасет. 
The ensemble has the most significant increase in the 1st test protocol with the largest training dataset.
% Наилучшие качество показывает ансамбль на основе ResNet50.
The ensemble based on ResNet50 has the best quality.
% Конструкция ансамбля позволяет снизить зависимость предсказания от границ изображения, и как следствие от выбора face bounding box. 
The ensemble design reduces the dependence on the image boundaries and, consequently, the face bounding box.
% Однако, независимое предсказывание ансамблем для каждого смещения существенно повышает вычислительные затраты. 
However, independent prediction for each offset significantly increases computational costs by 49 times. 
% However, independent prediction for each offset significantly increases computational costs.

\subsection{Knowledge distillation}

% Первоначально, мы решили использовать дистилляции знаний, чтобы сжать ансамбль до одной нейронной сети.
Initially, we decided to use KD to compress the ensemble into a single NN.
% Наиболее подходящим методом дистилляции знаний из ансамбля является response-based метод.
The most suitable method of KD from the ensemble is the response-based method.
% Другие методы дистилляции знаний из ансамбля выглядело трудоёмким без гарантированного результата.
Other KD methods from the ensemble are laborious without a guaranteed result.
% Наивный response-based метод  для задачи регрессии является неэффективным.
The naive response-based method for the regression problem is inefficient.
% Поэтому мы использовали RvC голову, которая сводит задачу регрессии к задаче классификации.
Therefore, we use the RvC head, which reduces the regression problem to the classification problem.
% Для классификации response-based методы ученик всегда показывает результат хуже учителя.
The student trained by the response-based method always has worse results than the teacher for the classification problem~\cite{Gou2020,10.1145/1150402.1150464, NIPS2014_ea8fcd92, hinton2015distilling}.
% Мы рассчитывали, что прирост от использования ансамбля будет нивелирован дистилляцией знаний.
We expected that KD would compensate for the ensemble gain.
% Однако, мы получили неожиданный результат в виде бустинга предсказаний сверточного ансамбля. 
However, we unexpectedly received a boost of convolutional ensemble predictions.

% Во всех тестовых протоколах в качестве учителя использовался сверточный ансамбль на базе ResNet50.
We use a convolutional ensemble based on ResNet50 as a teacher in all test protocols.
% Эффект бустинга проявился на 1ом и 2ом тестовых протоколах. 
There is the KD boosting effect on the 1st and 2nd test protocols.
% Однако, только на 1ом тестовом протоколе можно корректно сравнивать с предыдущими работами.
% However, we can correctly compare with previous works on the 1st test protocol.
However, we can correctly compare the 1st test protocol with previous works.
% KD-ResNet18 показывает результаты лучше ансамбля и уступает предыдущим работам только на датасете BIWI. 
KD-ResNet18 is inferior to previous works only on BIWI.
% При этом KD-ResNet18 имеет значительно меньше вычислительных ресурсов. 
At the same time, KD-ResNet18 requires significantly fewer computing resources.
% KD-ResNet50 уступает работе~cite{} только на датасете BIWI, но обходит все предыдущие работы на датасете AFLW2000.
KD-ResNet50 is inferior to~\cite{Valle2020}  only on BIWI but overtakes all previous work on  AFLW2000.
% Также, использование KD позволило обучить более глубокую модель KD-ResNet152 с существенным приростом точности.
Also, KD allows one to train deeper NN models with increased accuracy.
% Базовый ResNet152 отностильно ResNet50 показывал небольшой прирост точности на BIWI, но точность на AFLW2000 падала.
Base ResNet152 slightly improves the results of Base ResNet50 on BIWI but worsens the results on AFLW2000.
% KD-ResNet152 показывает прирост существенный точности на обоих датасетах.
KD-ResNet152 demonstrates a significant increase in accuracy on both datasets.

% Комбинация RvC голов, сверточного ансамбя и дистилляции знаний позволяет бустить точность HPE.
Proposed NN architecture, convolutional ensemble, and KD based on RvC improve the HPE accuracy.
% Предложенный метод может использоваться не только для сжатия нейронных сетей, но и в качестве метода обучения для получения нейронной сети с максимальной точностью.
The proposed method can be used not only to compress NNs but also as a training method to obtain NNs with maximum accuracy.
% Возможно, предложенный метод будет эффективен и для других задач регрессии.
Perhaps the proposed method will be effective for other regression problems as well.
}

\section{Conclusion}

% % Мы показали, что использование ансамбля смещений позволяет повысить точность вычисления углов поворота головы.
% We showed that the offset ensemble allows increasing the accuracy of head pose estimation.
% % Также мы показали, что применение подхода knowledge distillation для обучения нейронной сети на основе ансамбля смещений.
% We demonstrated that knowledge distillation (KD) allows increasing the accuracy of the offset ensemble.
% % Knowledge distillation улучшает результатов в среднем на 7.7\% по сравнению с вариантом без использования ансамбля.
% KD improves the results by an average of 7.7\% compared to the non-ensemble version.
% % Полученные результаты существенно превосходят или показывают сравнимые со state-of-the-art результатами.
% The results obtained either significantly exceed or are comparable with with state-of-the-art results.

% Мы предложили метод предобработки данных и архитектуру нейронной сети, содержащая 4 RvC головы и одну регрессионную голову. 
\optCyan{We propose a data preprocessing method and a NN architecture containing four RvC heads and one regression head.
% Эти улучшения позволили улучшить результаты предыдущих работ в некоторых тестовых протоколах.
These innovations improve the results of previous work in some test protocols.
ResNet50 trained on 300W-LP reduces MAE to 3.74 on AFLW2000.

% Мы предложили сверточный ансамбль, который снижает сглаживает выбросы и снижает зависимость от границ изображения. 
We propose the convolutional ensemble that reduces outliers and dependence on image boundaries.
The convolutional ensemble is a kind of test time augmentation.
% Ансамбль улучшенает качества на всех тестовых протоколах отностительно базовой нейронной сети.
The ensemble improves the quality of the base NN on all test protocols.
% It improves the quality of the base NN on all test protocols.

% Мы предложили response-based метод дистиляции знаний for HPE problem.
We propose the response-based method of KD for the HPE problem.
% В задачи классификации ученик всегдя хуже учителя, а для наивной регрессии response-based method бесполезен.
A student trained by the response-based method always has results worse than the teacher model for the classification problem~\cite{Gou2020,10.1145/1150402.1150464, NIPS2014_ea8fcd92, hinton2015distilling}.
For naive regression, the response-based method is useless.
% Ученик обученный предложенным методом показывает результаты лучше учителя, что не типично для response-based метода.
A student trained by the proposed method achieves results better than a teacher.
% Это неожиданная особенность позволяет использовать KD в качестве бустинга.
This unexpected feature makes it possible to use KD as a booster.
% Более того, метод позволил обучить более глубокую модель ResNet152, которая без KD имела точность схожую с ResNet50.
The proposed method provides an increased accuracy due to a deeper ResNet152, which without KD has an accuracy near ResNet50.
}
% Knowledge distillation улучшает результатов в среднем на 7.7\% по сравнению с вариантом без использования ансамбля.
KD improves the results by an average of 7.7\% compared to the non-ensemble version.
% Полученные результаты существенно превосходят или показывают сравнимые со state-of-the-art результатами.
The results obtained either significantly exceed or are comparable with the \optionalTwo{state-of-the-art}{SOTA} results.
% The proposed NN on the main test protocol improves the state-of-the-art result on AFLW2000 and approaches, with only a minimal gap, the state-of-the-art result on BIWI.
% Our NN uses only head pose data, but the previous state-of-the-art model also uses facial landmarks during training.

% Мы выложили в открытый доступ нейронные сети обученные на датасете 300W-LP.
We have made the trained NNs publicly available~\cite{angles_source}.
% ResNet152 показывает наилучшие результаты с точки зрения точности.
ResNet152 obtains the best results in terms of accuracy.
% ResNet18 показывает на AFLW2000 результат лучше, чем любой предыдущий метод. 
ResNet18 obtains a better result on \optionalTwo{the AFLW2000 dataset}{AFLW2000} than any previous method.
% ResNet18 может быть интересна для прикладных задачах, в которых важна максимальная производительность. 
ResNet18 may be of interest for application tasks where maximum performance is important.

% Также мы выложили в открытый доступ bounding boxes для датасетов 300W-LP, AFLW, AFLW2000 и BIWI.
We have made publicly available the face bounding boxes~\cite{angles_source} for \optional{the} 300W-LP, AFLW, AFLW2000, and BIWI\optional{ datasets}.
% Данная разметка позволяет выполнять более объективное сравнение алгоритмов и нивелировать погрешности методов связанные с выбором bounding boxes.
This labeling allows one to perform a more objective comparison of methods and level out the errors associated with the choice of the bounding boxes.

% or


% \appendices

% \section{}

% \ifCLASSOPTIONcompsoc
%   % The Computer Society usually uses the plural form
%   \section*{Acknowledgments}
% \else
%   % regular IEEE prefers the singular form
% \fi


\ifCLASSOPTIONcaptionsoff
  \newpage
\fi



%\bibliographystyle{IEEEtran}

% H.~Kopka and P.~W. Daly, \emph{A Guide to {\LaTeX}}, 3rd~ed.\hskip 1em plus

% \end{thebibliography}

% \bibliographystyle{IEEEtran}
\bibliography{IEEEabrv,mybibliography}


% \includegraphics[trim=left bottom right top, clip]{file}
\begin{IEEEbiography}[{\includegraphics[trim=12 5 28 35, width=1in,height=1.25in,clip,keepaspectratio]{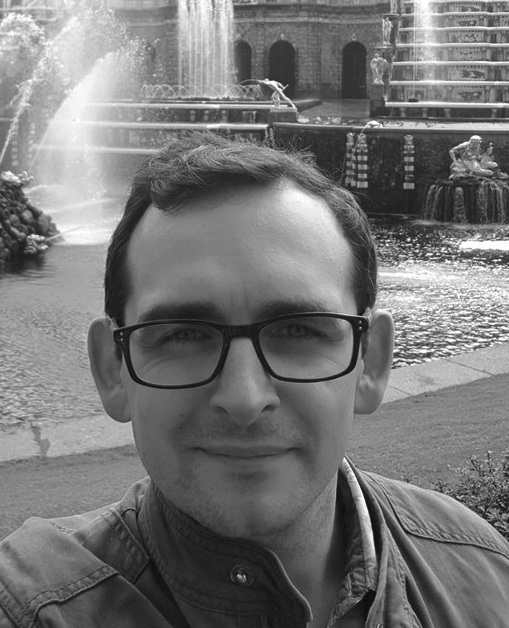}}]{Andrey Sheka}
A. S. Sheka is a researcher in the Laboratory of Complex Systems Analysis of the Computational Systems Department at the N.N. Krasovskii Institute of Mathematics and Mechanics. 
His main research interests are Deep Learning, Computer Vision and Behavioral Analysis.
He has published over 30 publications on data analysis, artificial intelligence, and computational optimization.
\end{IEEEbiography}

\begin{IEEEbiography}[{\includegraphics[trim=0 0 0 0,width=1in,height=1.25in,clip,keepaspectratio]{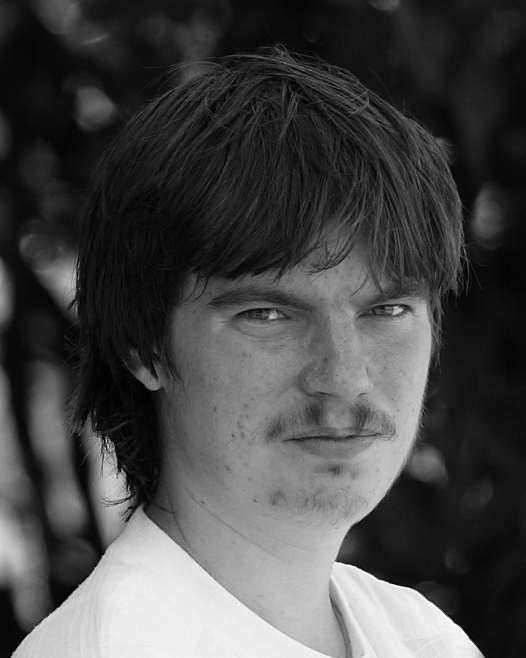}}]{Victor Samun}
V. S. Samun is a researcher in the Laboratory of Complex Systems Analysis of the Computational Systems Department at the N.N. Krasovskii Institute of Mathematics and Mechanics.
His main research interests are Deep Learning and Computer Vision.
He has published 5 publications on high-performance computing, machine learning, and artificial intelligence.
\end{IEEEbiography}

\vfill


%\newpage


%\vfill


\end{document}